\documentclass[10pt]{article} 
\usepackage[preprint]{tmlr}

\usepackage{amsthm}
\newtheorem{proposition}{Proposition}
\usepackage{placeins}
\usepackage{booktabs}
\usepackage{graphicx}
\usepackage{color, soul}
\usepackage{tabularx}
\sethlcolor{yellow}
\usepackage{subcaption}
\usepackage{booktabs}
\usepackage{tabularx}
\usepackage{caption}
\usepackage{hyperref}
\usepackage{amsmath}
\usepackage{multirow}
\usepackage{caption}
\usepackage{amsthm}
\usepackage{longtable}
\usepackage{amsmath,amssymb}
\usepackage{algorithm}
\usepackage{eurosym}
\usepackage{algpseudocode}
\usepackage{eurosym}
\newtheorem{lemma}{Lemma}

\usepackage{hyperref}
\usepackage{url}

\title{Local Matters: Calibration Requirements in Online Conformal Forecasting}

\author{\name Manush Sanchela \email manush\_sanchela@dau.ac.in   \\
      \addr Dhirubhai Ambani University, Gandhinagar.
      \AND
      \name Pritam Anand  \email  pritam\_anand@dau.ac.in \\
      \addr Dhirubhai Ambani University, Gandhinagar.
      }

\def\month{MM}  
\def\year{YYYY} 
\def\openreview{\url{https://openreview.net/forum?id=XXXX}} 

\begin{document}

\maketitle

 \begin{abstract}
This paper emphasizes the significance of the Local Calibration Requirement
 (LCR) for probabilistic forecasting models used in real-time, risk-aware
 decision support. We first formalise different granularities of calibration
 requirements for probabilistic forecasting models and analyze existing models
 through the lens of these requirements. We show that, while Adaptive
 Conformal Inference (ACI) meets the global calibration requirement, it
 satisfies the LCR only to a limited extent, particularly under sudden
 distribution shifts. However, operational decisions are made
over short time local operational windows, during which Prediction Intervals (PIs) may fail repeatedly even when the long-run coverage is on target. It necessitates the adoption of  local evaluation metrics that
quantify how often, and how severely, PIs fail within operating windows. Motivated by this analysis, we propose Locally Adaptive
 Conformal Inference (LACI), a simple modification of the ACI recursion that
 incorporates local coverage errors into its weighting mechanism to improve
 local calibration. We derive theoretical results showing that, under abrupt
 changes in the data distribution, LACI applies a stronger corrective update
 than ACI and hence can
improve the calibration within operational local window. Extensive experiments on nine real-world energy datasets show that
 LACI improves the local calibration of ACI and its variants, along with a case study that
 shows that this improvement translates into lower operational costs in energy application.
\end{abstract}

\section{Introduction}
Anticipating the future has long been central to human planning, and
forecasting models now underpin decisions in domains as diverse as energy,
finance, healthcare, climate and agriculture by turning historical data into actionable
predictions. The evolution of modern deep sequential architectures has improved the quality of the data-driven forecast that in turns has improved the quality of decisions across diverse domain. Such improvements often leads significant economical and social impacts. For instance, \citet{jeon2022estimating} estimate that two upgrades of NOAA's
operational wind forecasts could have saved US electricity consumers
approximately \$384 million through lower electricity prices.

Given the time series observations $x_1, x_2, \ldots, x_t$, forecasting models
aim to estimate future observations through the historical sequential
auto-regressive relationship. However, such relationships are often stochastic,
non-stationary and chaotic, which limits the prediction quality of forecasting
models irrespective of the complexity and sophistication of their design, and
hence limits their usability in final decision-making.

In view of the above limitation of the point forecasting models, researchers have shifted their attention to the probabilistic forecasting models in recent years for risk-aware decision support.  This shift is now
evident across many domains: in energy systems for wind power
\citep{xie2023overview}, electrical load \citep{vandermeer2018review,hong2016probabilistic} and wave
height \citep{taylor2018probabilistic}; in finance for market volatility \citep{eggen2026financial} and risk measures prediction \citep{nieto2016frontiers}; in healthcare for epidemic planning
\citep{rodriguez2024machine}; and in retail demand
planning \citep{makridakis2022m5}. Probabilistic forecasting models quantify the uncertainty in their predictions
by estimating a range of values within which future observations
are expected to fall with a prescribed high probability (say $95\%$) that makes the forecasts more actionable and trustworthy for
critical real-world decisions.

From the statistical perspective, the problem of probabilistic forecasting is fundamentally a Prediction Interval (PI) estimation problem with sequential data. It makes the probabilistic forecasting algorithms naturally move around the quantile estimation methods and evolve in two major directions across the literature. The first assumes an underlying parametric distribution and estimates its parameters, from which pair of quantile functions desired for the PI is derived; representative approaches include Normalizing Flows \citep{rezende2015variational}, Mixture Density Networks \citep{bishop1994mixture}, and DeepAR \citep{salinas2020deepar}. The second  direction estimates the desired quantile functions in distribution-free setting and often directly by minimizing the pinball loss \citep{koenker1978regression} using deep sequential architectures. However, compared with distribution-based algorithms, pinball-loss-based  learning
models, often termed with quantile-based learning models, have become the
dominant choice in practice owing to their consistent performance across a
wide variety of datasets and applications.

Even for cross-sectional data, PI estimation models often fail to guarantee the
target coverage, particularly when the training sample is small. Conformal
regression \citep{vovk2005algorithmic} provides a promising post-hoc procedure
that adjusts the PI to guarantee the target coverage in finite samples under
the assumption of exchangeability. However, conformal regression cannot be
directly applied to probabilistic forecasting with the same guarantees, since
the sequential nature of time series data inherently violates the
exchangeability assumption. \citet{gibbs2021adaptive} addressed this gap by
introducing Adaptive Conformal Inference (ACI), which guarantees that the
long-run miscoverage rate converges to the target miscoverage level, with an explicit
finite-sample bound, even under arbitrary shifts in the data distribution.

In
ACI, the miscoverage level is updated online through a simple recursion driven
by the observed coverage errors, which makes it particularly effective for
improving the quality of PIs produced by probabilistic forecasting models in
online settings. The ACI framework therefore offers a highly promising pathway that operates on top of any probabilistic forecasting model to improve the reliability of its
PIs across different application domains. Accordingly, the recent literature
has started to observe the increasing use of the ACI framework in diverse applications, such as wind power forecasting \citep{li2026spatio}, electricity price forecasting \citep{suresh2025mamba,o2025conformal}, solar power forecasting \citep{suresh2026non} and marker risk management \citep{fantazzini2024adaptive,lima2026conformal}. 
\begin{figure}[t]
    \centering
\subfloat[]{
        \includegraphics[width=0.8\linewidth]{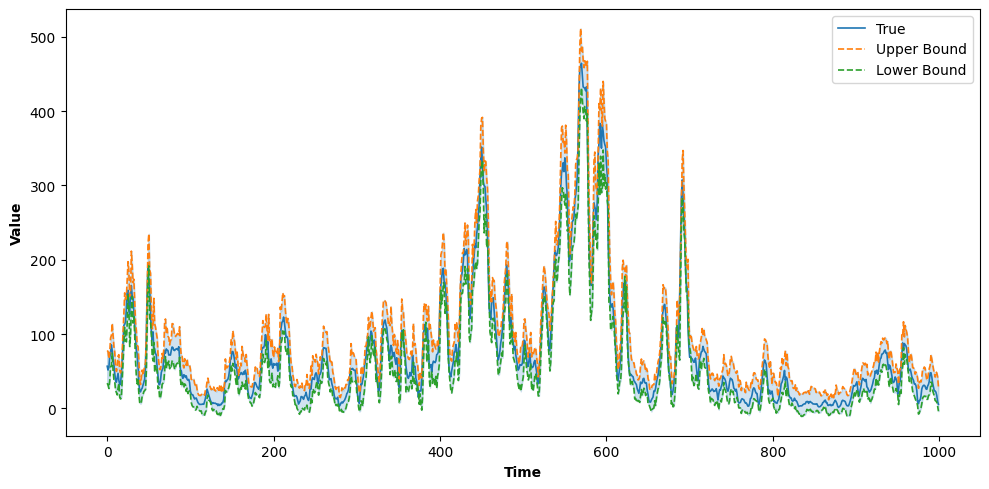} 
    }    \\
    \subfloat[]{
        \includegraphics[width=0.8\linewidth]{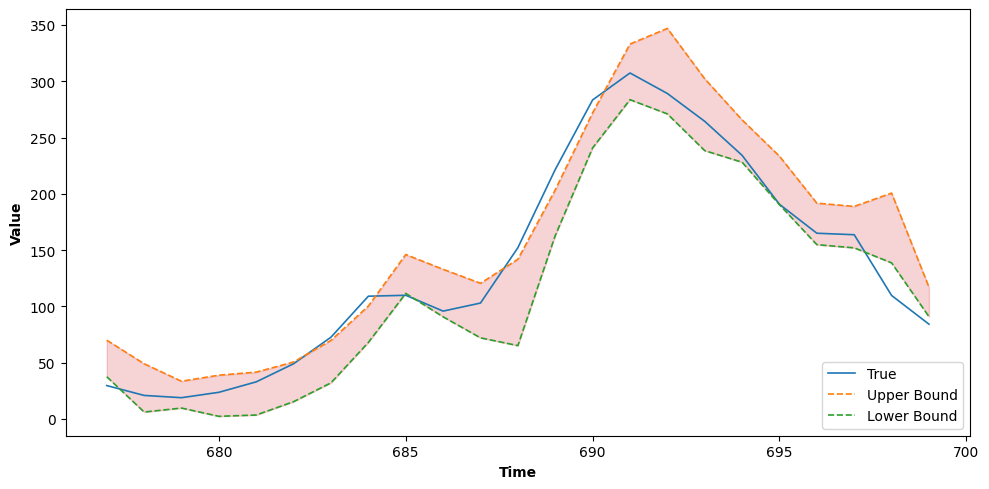}
    }
    
      \caption{\footnotesize{
    (a) Rolling wind power forecasts of the LSTM-Quantile model over 1000 future time steps. The model achieves a global PICP of \textbf{90.2}\%, which suggests overall reliability. However, it fails to maintain the required level of local calibration, which is essential for risk-aware daily decision-making in grid operations. 
    (b) Local performance of the LSTM-Quantile model over selected 24-hour windows. Only \textbf{54.7}\% of the observations fall within the estimated prediction intervals in these windows, highlighting poor local calibration and reducing the practical usefulness of the forecasts.
    }}
    \label{fig:local_plot}
\end{figure}

Despite the evolution of effective deep probabilistic forecasting models and
conformal calibration methods, two research gaps remain evident. First, it is
unclear how probabilistic forecasts should be translated explicitly into
actions for risk-aware decision support. Second, it is unclear how these
models should be evaluated so as to reflect their actual usefulness in
risk-aware decision-making. At present, most of them are typically evaluated only
through global metrics such as the Prediction Interval Coverage Probability
(PICP) and the Mean Prediction Interval Width (MPIW), which average over the
entire test period. Such global metrics can mask periods in which the PIs
repeatedly fail to cover the observations, even when the overall coverage is
close to its target, and it is precisely during these periods that decisions
based on the PIs are most likely to go wrong. This is well illustrated through a use case in Figure \ref{fig:local_plot}.


We argue that, while the solution to the first gap may be application
dependent, the value of probabilistic forecasting models that perform well
under local evaluation is application agnostic, in any domain, decisions are
made over finite operating windows, and a model that fails repeatedly within
such windows is of limited practical use, however good its overall coverage.

The need for locally reliable PIs in the online time series setting has only
recently begun to receive attention. Notable efforts in this direction are
DtACI \citep{gibbs2024conformal} and SAOCP \citep{bhatnagar2023improved}.
\citet{gibbs2024conformal} develop a variant of ACI that aggregates multiple
ACI updates with different step sizes, so as to perform well over arbitrary
time intervals rather than only over the entire horizon, while
\citet{bhatnagar2023improved} track the interval radius directly in the score
space by aggregating online learners started at different times. However,
these efforts address local reliability only indirectly by complicating the overall on-line learning process through the mixture of experts, as  their
guarantees are stated in terms of the regret of the pinball loss,
and not in terms of the coverage attained within local windows. In
operational settings such as energy system management, operators are
concerned with how often, and how severely, the PIs fail across local
operating windows.

Moreover, there remains a significant gap in the literature in
understanding the calibration of probabilistic forecasting models at
different granularities. The forecasting literature lacks a clear definition
of these granularities in the online setting, principled metrics to evaluate
each of them, and an understanding of where existing probabilistic
forecasting solutions, such as quantile-based models, conformal methods, and
ACI and its variants, lie with respect to them. Addressing this gap is
essential for identifying which solutions are reliable enough to support
risk-aware decisions in practice. The contributions of this paper are as
follows.

\begin{enumerate}
  \item We formalise different granularities of calibration requirements for probabilistic forecasting models, and highlight the significance of the Local Calibration Requirement (LCR) for decisions made over operating windows. We further propose evaluation metrics that does not only quantify the quality of local calibration but, also  directly participate into the risk-aware energy decisions.

  \item We analyze existing probabilistic forecasting models through the lens
  of these requirements. We show that a Quantile Regression (QR) model does
  not consistently meet even the Global Calibration Requirement (GCR), while ACI meets the GCR but satisfies the LCR only to a limited extent. In
  particular, we derive an exact identity linking the local coverage error of
  ACI to the drift of its miscoverage level within the local operational window. 

  \item We carefully study the  local coverage error trajectory generated by ACI and identify an inequality that characterizes the ACI process in terms of local coverage errors well in case of the abrupt changes in auto-regressive distribution. The ACI process often yields abrupt local coverage errors, particularly in the case of significant distribution shift. To enhance the usability of ACI for real-time  risk-aware decision support, we realize  that its local coverage errors should remain small even under significant
distribution shifts.

  \item In view of the above limitation of the ACI, we develop  Local Adaptive Conformal Inference
(LACI), a simple but effective variant of the ACI, that  weights past errors by the local
coverage error, thereby penalizing abrupt changes in local coverage error more heavily and produce a smooth local error trajectory. We provide theoretical support for this design, showing that,
from a common state and under a strict persistence condition on local coverage failures, the
LACI process must improves the local coverage error relative to
ACI. Further, we empirically show upon nine energy related time-series datasets that the use of the LACI  leads the significant improvement in the local calibration ability and this improvement leads to substantial monetary saving in energy applications.
 \end{enumerate}

The rest of the paper is organised as follows.
Section~2 reviews the background required for the
subsequent sections. Section~3 formally establishes the different
granularities of calibration and develops local evaluation metrics that enter
directly into the end decision functions. Section~4 analyses
existing forecasting methods from the lens of established calibration requirements and also highlights the limitations and advantages of the ACI algorithm.
Section~5 introduces the proposed Local Adaptive Conformal
Inference (LACI) method along with a few  supporting theoretical results. Section~6 presents empirical
results supporting the claims made in this paper, and
Section~7 discusses its limitations and outlines directions
for future research.
 
\section{Related Work}
This section systematically reviews the existing UQ concepts that builds the foundation of the modern on-line probabilistic forecasting models.

\subsection{Prediction Interval Estimation}
Given training dataset $\{(x_i, y_i) : x_i \in \mathbf{R}^d,\, y_i \in \mathbf{R},\, i = 1,2,\dots,n \}$, consisting of independent and identically distributed samples drawn from the joint distribution of the random variables $(X, Y)$, the PI estimate with target coverage $1-\alpha$, is a pair of functions $[f_1,f_2]$ that must satisfy at least the global calibration requirement  given by
\[
P\big( f_1(x) \leq Y \leq f_2(x) \big) \geq 1 - \alpha.
\]
Although multiple PIs may satisfy this coverage requirement, narrower intervals are generally preferable, as they provide more precise and informative predictions. Thus, the objective is to achieve the desired coverage while minimizing the average prediction interval width. The two bounds of a PI can be viewed as a pair of conditional quantile functions, estimating the \(q\)-th and \((1-\alpha+q)\)-th quantiles of the conditional distribution \(Y\mid X\), respectively, for some \(q\in(0,\alpha)\).

\subsection{Conformal Regression}
A well-calibrated PI may fail to achieve the desired coverage in finite samples, particularly when the underlying quantile functions are estimated from limited data. Moreover, standard quantile regression does not provide a distribution-free finite-sample coverage guarantee. Conformal Regression(CR) \cite{vovk2005algorithmic} addresses this limitation by applying a post-hoc calibration step to the estimated PIs. By using a separate calibration dataset, it adjusts the interval bounds to provide a finite-sample marginal coverage guarantee under the exchangeability assumption, while retaining the predictive information captured by the underlying regression model.

Let $\mathcal{D}_{\mathrm{train}}$ and $\mathcal{D}_{\mathrm{cal}}$ denote the training and calibration datasets, respectively, with $m$ observations in the calibration set. First, a regression function $\widehat f$ is estimated using $\mathcal{D}_{\mathrm{train}}$. For each calibration observation $(x_i,y_i)\in\mathcal{D}_{\mathrm{cal}}$, the nonconformity score is computed using the absolute residual:
\[
R_i = |y_i-\widehat f(x_i)|,
\qquad i=1,\ldots,m.
\]
These scores quantify the prediction errors on the calibration set. Alternatively, when the initial PI $[\widehat f_1,\widehat f_2]$ is constructed , the nonconformity score of \cite{romano2019conformalized} is given by
\[
R_i =
\max\left\{
\widehat f_1(x_i)-y_i,\;
y_i-\widehat f_2(x_i)
\right\},
\qquad i=1,\ldots,m.
\]
The conformal adjustment requires the computation of the $Q_{1-\alpha} = (\left\lceil (m+1)(1-\alpha)\right\rceil)^{th}$ quantile of $R_i$ .
For a new test observation with covariates $x_{m+1}$, the standard residual-based split conformal prediction set is constructed as
\[
\widehat C^{(\alpha)}(x_{m+1})
=
\left[
\widehat f(x_{m+1})-\widehat Q_{1-\alpha},
\;
\widehat f(x_{m+1})+\widehat Q_{1-\alpha}
\right].
\]
Similarly, using the nonconformity score of \cite{romano2019conformalized}, the conformalized prediction set is given by
\begin{equation}
\widehat C^{(\alpha)}(x_{m+1})
=
\left[
\widehat f_1(x_{m+1})-\widehat Q_{1-\alpha},
\;
\widehat f_2(x_{m+1})+\widehat Q_{1-\alpha}
\right].
\label{eq:conformity_score}
\end{equation}

\subsection{Probabilistic Forecasting}

Let us consider time-series observations 
$x_1, x_2, \ldots, x_t$ as realizations of the stochastic process 
$\{X_j\}_{j=1}^{t}$. The objective of the forecasting task  is to reveal the auto-regressive relationship by characterizing the conditional 
distribution of the future value $X_{t+1}$ given the available historical 
information $H_t$,  where 
\[
H_t = \{X_t, X_{t-1}, \ldots, X_{t-w+1}\}
\]
denotes the collection of the most recent $w$ variables. Conventional forecasting models primarily focus on estimating the conditional expectation $\mathbf{E}[X_{t+1} \mid H_t]$, which provides a point prediction of the future value. For the risk-aware decision function, the probabilistic forecasting models with target coverage $1-\alpha$ construct the PI by estimating an appropriate pair of quantile functions of $X_{t+1}|H_t$ such that it covers the $1-\alpha$ fraction of the observations by sacrificing the minimum possible width.

\subsection {Probabilistic Forecasting metrics}
We detail the existing evaluation metrics that are used to evaluate the quality of the probabilistic forecast. We also highlight how these metrics are often insufficient to support the risk-aware energy decisions due to their inability to reflect the local calibration ability of the trained model.   

In literature, the probabilistic forecasting models are commonly trained upon the time-series  data $\{x_1,x_2,\ldots x_t \}$ and evaluated upon the test data $\{x_{t+1},x_{t+2},\ldots x_{t+k}\}$  with the following metrics.

\begin{enumerate}
\item[(a)] \textbf{PICP (Prediction Interval Coverage Probability)}:- It measures the fraction of true observations that fall within the PI over the test horizon. It is defined as:

\[
\mathbf{PICP} = \frac{1}{k} \sum_{i=1}^{k} \mathbb{I}
\left( x_{t+i} \in [\hat{L}(x_{t+i}), \hat{U}(x_{t+i})] \right)
\]

where $[\hat{L}(x_j), \hat{U}(x_j)]$ is the estimated PI for $j^{th}$ time-stamp.
For a well-calibrated model, the PICP should be close to or greater than the nominal confidence level $1-\alpha$.

\item[(b)] \textbf{MPIW (Mean Prediction Interval Width)} :- MPIW quantifies the average width of the estimated PI and is given by:
\begin{equation*} \mathbf{MPIW} = \frac{1}{k} \sum_{i=1}^{k} \left( \hat{U}(x_{t+i}) - \hat{L}(x_{t+i}) \right) \end{equation*}
Lower MPIW values correspond to sharper PIs that are more actionable in real-time decision support.
\item[(c)] \textbf{CRPS (Continuous Ranked Probability Score)} :-CRPS evaluates the quality of the full predictive distribution. However, it is primarily relevant for models that explicitly estimate the conditional autoregressive density and subsequently derive the quantile function from it, such as DeepAR  and Mixture Density Network (MDN)-based models . Also, it rarely participates in the final decision function for risk-aware decision support.

\end{enumerate}

\section{ Calibration Levels: Definitions and Metrics}

Motivated by the limitations of global calibration metrics, we aim to develop
a local calibration framework that enters directly into real-world, risk-aware
decision support. To this end, we first clarify what calibration means for PI
estimation, and then formally establish the different granularities of the
calibration requirement for probabilistic forecasts.

\subsection{Calibration Levels for PI estimation}

At the outset, we revisit calibration requirements across different levels of granularity for PI estimation. In this context,  \cite{chung2021beyond} introduce an oracle framework that characterizes these varying calibration notions for quantile estimation in cross-sectional data, while also highlighting the inherent limitations of existing quantile estimators.

Following the framework of \cite{chung2021beyond}, we begin by formalizing different levels of calibration requirements expectation from the estimated PI $[f_1, f_2]$ over the training dataset $\{(x_i, y_i) : x_i \in \mathbf{R}^d,\, y_i \in \mathbf{R},\, i = 1,2,\dots,n \}$, consisting of independent and identically distributed samples drawn from the joint distribution of the random variables $(X, Y)$ as follows.

\begin{enumerate}
    \item [(a)] \textbf{Individual Calibration}:- The estimated PI $[f_1, f_2]$ is said to be well-calibrated at a point $x$ if it satisfies
\[
P_{Y \mid x}\big( f_1(x) \leq Y \leq f_2(x) \big) \geq 1 - \alpha.
\]
Furthermore, the PI $[f_1, f_2]$ is said to be \textit{individually calibrated} if
\[
P_{Y \mid  x}\big( f_1(x) \leq Y \leq f_2(x) \big) \geq 1 - \alpha, \quad \forall x \in \mathcal{X}.
\]
In other words, for each $x \in \mathcal{X}$, the probability that the corresponding response $Y$ lies within the interval $[f_1(x), f_2(x)]$ must be at least $1 - \alpha$. 

    \item [(b)] \textbf{Global Calibration:-} The estimated PI $[f_1, f_2]$ is said to be (\textit{globally}) well-calibrated if it satisfies
\[
E_{X}\big( P_{Y|X}\big( f_1(X) \leq Y \leq f_2(X) \big)\big)\geq 1 - \alpha.
\]
It means that, on average, the probability that \( Y \) lies within the PI \( [f_1(X), f_2(X)] \) is at least $1 - \alpha$. The \textbf{PICP} metric typically empirically measures the quality of the global calibration for the given sample set.

    \item [(c)] \textbf{Local Calibration:-} The estimated PI $[f_1, f_2]$ is said to be \textit{locally well-calibrated} around a point $x \in \mathcal{X} $ if it is well-calibrated within a neighborhood of $x$. Let $\|\cdot\|$ denote a norm on $\mathbf{R}^n$ and let $\epsilon > 0$ be a fixed radius then  define the $\epsilon$-neighborhood of $x$ as
\[
\mathcal{N}_\epsilon(x) = \{x' \in \mathcal{X} : \|x' - x\| \leq \epsilon\}.
\]
Let $X_\epsilon$ denote the random variable conditioned upon the event $X \in \mathcal{N}_\epsilon(x)$. Then, the PI $[f_1, f_2]$ is locally calibrated around $x$ if
\[
E_{X_\epsilon}\big( P_{Y|X}\big( f_1(X) \leq Y \leq f_2(X) \big)\big)\geq 1 - \alpha.
\]
\end{enumerate}
We note that individual (pointwise) calibration is a stringent requirement and, in general, cannot be guaranteed from finite samples without imposing additional assumptions. In contrast, globally calibrated PIs, while easier to achieve, may be inadequate in many decision-making settings due to their inability to capture localized uncertainty and support risk-aware decisions at a finer granularity. This necessitates the development of UQ algorithms capable of constructing PIs that are well-calibrated at a local level. However, apart from such calibration requirements, the width of the PI should be expected to be as small as possible for reducing the overall uncertainty in the resulting decision.

\subsection{Calibration Levels for Probabilistic Forecasting}
Returning to our auto-regressive forecasting objective that characterizes the conditional 
distribution of the future value $X_{t+1}$ given the available historical 
information $H_t$,  where 
\[
H_t = \{X_t, X_{t-1}, \ldots, X_{t-w+1}\}
\]
denotes the collection of the most recent $w$ variables. Conventional forecasting models primarily focus on estimating the conditional expectation $\mathbf{E}[X_{t+1} \mid H_t]$, which provides a point prediction of the future value. For the risk-aware decision function, the probabilistic forecasting models construct the PI by estimating an appropriate pair of quantile functions of $X_{t+1}|H_t$. However, we highlight the following concerns regarding the calibration requirements of PIs in auto-regressive forecasting setting.

\begin{enumerate}
   \item[(a)] The \textbf{Individual Calibration Requirement} (ICR) seeks the \textbf{forecast PI} $[f_1, f_2]$ such that
${P}_{X_{i+1}\mid h_i}\big( f_1(h_i) \leq X_{i+1} \leq f_2(h_i) \big) \geq 1 - \alpha,  \forall\, h_i \in \mathcal{H} ~~i \geq w$. However, in forecasting settings, it is generally infeasible to construct PIs that achieve exact individual calibration at the target coverage level of $1 - \alpha$. This limitation arises because, for any given $h_i$, $ i \geq w$, only a single realization of the conditional random variable $X_{i+1}\mid h_i$ is observed in the time-series data. Consequently, the empirical individual coverage of $h_i$ for the estimated PIs oscillates between $0$ and $1$. Let us denote this empirical individual coverage for $h_i$ with $z_i$ in our future discussion.

\item [(b)]  For $i\geq p$, the \textbf{Global Calibration Requirement} (GCR) enforces the \textbf{forecast PI} $[f_1, f_2]$ must satisfy
\begin{equation*}
E_{H_i}\big( P_{X_{i+1}|H_i}\big( f_1(H_i) \leq X_{i+1} \leq f_2(H_i) \big)\big)
\geq 1 - \alpha
\end{equation*}
that simplifies to
\begin{equation*}
P\big( f_1(H_i) \leq X_{i+1} \leq f_2(H_i) \big)
\geq 1 - \alpha, \quad i \geq w.
\end{equation*}
It only requires that on average, the probability that time-series observations $X_{i+1},i\geq w$,  would lie inside the forecast PI $[f_1(H_i), f_2(H_i)]$ should be at least $1-\alpha$. However, solely ensuring the \textbf{global calibration requirement} in forecasting tasks may provide a loose calibration requirement for the PI in many time-series applications, particularly when working with long test horizons. It can often render the PI non-actionable in most cases. Figure \ref{fig:local_plot} presents a usual scenario in wind power probabilistic forecasting where the forecast appears to be well-calibrated \textit{globally} but may not be actionable in real-world daily operations.
    
\end{enumerate}

\subsubsection{Local Calibration requirement for the forecast PI}

In view of the insufficiency of the global calibration requirement and the impracticality of the individual calibration requirement in probabilistic forecasting tasks, we grow the idea of local calibration requirement systematically in time-series uncertainty forecast to make the forecasts more actionable in real-world decision support.

For this, we first define three distinct definition of the neighborhood of $h_i$, $i \geq w$, with the window $w$ using 
\begin{eqnarray}
    \mathcal{N}_w^{L}(h_i) = \{ h_j \mid j = i-w+1, \ldots, i \}, ~~~~~~~~~~
    \mathcal{N}_w^{R}(h_i) = \{ h_j \mid j = i, \ldots, i+w-1 \} \nonumber \\
    \mbox{and }  \mathcal{N}_w^{C}(h_i) = \{h_j | j= i-l,\ldots, i+r \} \mbox{~where~} 
    l =
    \begin{cases}
        \frac{w}{2} - 1, & \text{if $w$ is even}, \\
        \left\lfloor \frac{w}{2} \right\rfloor, & \text{if $w$ is odd},
    \end{cases} \mbox{and} 
    \quad
    r = \left\lfloor \frac{w}{2} \right\rfloor l
    \label{neigh}
\end{eqnarray}

If we define a random variable $H_w^{(i)}$ that is conditioned upon the event $H \in \mathcal{N}^{I}_w(h_i)$, for $I \in \{L,C,R\}$ then the local calibration requirement for estimated forecast PI $\left[ f_1, f_2 \right]$ can be stated as follows.

The estimated forecast PI $[f_1, f_2]$ is said to be \textit{locally well-calibrated} at $i^{th}$ time-stamp, $i\geq p$, if    
\begin{equation}
E_{H_w^{(i)}} \big( P_{X_{i+1}|H_i}\big( f_1(H_i) \leq X_{i+1} \leq f_2(H_i) \big)  \big) \geq 1 - \alpha,
    \label{local_for}
\end{equation}
that is equivalent to
\begin{equation}
 P{\big( f_1(H_i) \leq X_{i+1} \leq f_2(H_i) | {H_w^{(i)}}\big)} \geq 1 - \alpha.
  \label{local_for1}
\end{equation}

The \textbf{Local Calibration Requirement} (LCR) (\ref{local_for1}) for the \textbf{forecast PI} is stricter than the GCR. However, it loosens with the growth of the size of the local window $w$ and asymptotically coincides with the GCR for sufficiently large $w$. The choice of the local window size $w$ in the LCR is guided by the needs of the forecasting application. For instance, to support risk-aware daily decision-making in a power grid, the window size $w$ may be set to 24 hours.

Another important point concerns the choice of the local calibration requirement, i.e., \textit{Left}, \textit{Right}, or \textit{Center}, which depends on the neighborhood defined in~\ref{neigh}. This choice depends on the context and the requirements of the application. However, when training the learning model to obtain locally calibrated probabilistic forecasts, it is more natural to optimize the objective using metrics computed over a local window built from the \textit{Left} neighborhood definition. In contrast, when evaluating the probabilistic forecasting model, the quality of local calibration may also be measured using the \textit{Center} or \textit{Right} neighborhood definition for supporting the risk-aware decisions.

\subsubsection{ Evaluation Metrics for the Local Calibration}
\label{def_local_met}
Now, we develop an evaluation metric that empirically quantifies the quality of LCR. We propose the following metric to measure the local calibration quality of forecast PI  $[f_1, f_2]$ upon the future test observations $x_{t+1}, x_{t+2}, \ldots, x_{t+k}$.






\begin{enumerate}
    \item [(a)] The local coverage at index $i$ is given by
    \begin{equation}
    \mathbf{LPICP}_i = \frac{1}{w} \sum_{j \in \mathcal{N}^{C}_w(h_i)} 
    \mathbb{I}\big[f_1({h_j}) \le x_{j+1} \le f_2({h_j}) \big],
    \end{equation}
 and, the overall average local coverage is computed as
    \begin{equation}
    \mathbf{LPICP} = \frac{1}{k} \sum_{i=t+1}^{t+k} \mathbf{LPICP}_i,
    \label{123}
    \end{equation}
    that is different from the commonly used PICP.

The \textbf{Left-Centered Local Coverage} at index $i$ is may also be given by
\begin{equation}
    \mathbf{LPICP}_i^{L} = \frac{1}{w} \sum_{j \in \mathcal{N}_w^{L}(h_i)}
    \mathbb{I}\big[ f_1(h_j) \le x_{j+1} \le f_2(h_j) \big].
    \label{lpicp}
\end{equation}

    For making risk-aware decisions with confidence level $1-\alpha$, each observed $\mathbf{LPICP}_i$ should ideally satisfy $\mathbf{LPICP}_i \ge 1-\alpha$ for $i = t+1, \ldots, t+k$. This is a significantly stronger requirement than ensuring that the overall PICP exceeds $1-\alpha$. 

    \item[(b)]   \textbf{Local Prediction Interval Coverage Error (LPICE):-} The LCR is violated whenever $\mathbf{LPICP}_i$ falls below the target coverage level $1 - \alpha$ for any index $i$, resulting in a Local Error (LE) that can be quantified at the $i^{\text{th}}$ index as 
    \begin{equation}
    \mathbf{LE_i} = \max\big((1 - \alpha) - \mathbf{LPICP}_i,\, 0\big)
    \end{equation}

The average of these $\mathbf{LE_i}$ can be termed as \textbf{Local Prediction Interval Coverage Error (LPICE)}  and given by
\begin{equation}
\mathbf{LPICE} = \frac{1}{k} \sum_{i=t+1}^{t+k} 
\max\big( (1-\alpha) - \mathbf{LPICP}_i, \, 0 \big).
\end{equation}

    The Left-Cenetered Local Error at index i is given by : 
    \begin{equation}
    \mathbf{LE_i^{L}} = \max\big((1 - \alpha) - \mathbf{LPICP}_i^{L},\, 0\big), \label{le}
    \end{equation}
and accordingly Left-Cenetered  LPICE can be derived.

A more practical approach to quantify LCR violation is through \textbf{LPICE} and the variance of $\mathbf{LE_i}$, which should be centered around zero for a locally well-calibrated forecast PI.

\item [(c)] \textbf{Local Prediction Interval Accuracy (LPIA):-}  The accuracy of the LCR can be defined as
\begin{equation}
\mathbf{LPIA} = { \frac{1}{k} \sum_{i=t+1}^{t+k} \mathbb{I}\big[\mathbf{LPICP}_i \geq 1-\alpha\big]}.
\end{equation}

The $\mathbf{LPIA}$ measures the proportion of indices at which the estimated forecast PI achieves local calibration.

\item[(e)] \textbf{Local Mean Prediction Interval Width (LMPIW) }:- To quantify the amount of the uncertainty at $i^{th}$ index locally for the forecast PI $[f_1,f_2]$, we define the \textbf{LMPIW} as
\begin{equation}
    \mathbf{LMPIW}_i = \frac{1}{w} \sum_{j \in \mathcal{N}^{C}_w(h_i)} 
    \big|f_2({h_j}) -f_1({h_j}) \big|,
    \end{equation}

and, the overall average local width is computed as
    \begin{equation}
    \mathbf{LMPIW} = \frac{1}{k} \sum_{i=t+1}^{t+k} \mathbf{LMPIW}_i,
    \label{12311}
    \end{equation}
    that is different from the commonly used MPIW. The \textbf{LMPIW} quantifies the local \textit{sharpness} of the forecast PI $[f_1,f_2]$.

\item[(f)] \textbf{Local Coverage Width-based Criterion (LCWC)}:- 
To quantify the local trade-off between \textit{coverage} and \textit{sharpness} of the forecast PI $[f_1,f_2]$ at the $i^{\text{th}}$ index, we define the local analogue of the CWC metric \citep{khosravi2010lower}  as follows.
\begin{equation}
\mathbf{LCWC}_i
=
\mathbf{NLMPIW}_i
\left(
1
+
\gamma_i
e^{
\left(
-\eta
\left(
\mathbf{LPICP}_i - (1-\alpha)
\right) \right)
}\right).
\end{equation}
Here NLMPIW stands for Normalized Local Mean Prediction Interval (NLMPIW) that is computed by
\[
\mathbf{NLMPIW}_i
=
\frac{\mathbf{LMPIW}_i}{R_i},
\]
where

\[ R_i
=
\max_{j \in \mathcal{N}^{C}_w(h_i)} x_j
-
\min_{j \in \mathcal{N}^{C}_w(h_i)} x_j \]
Also,  $
\gamma_i
=
\mathbb{I}
\left\{
\mathbf{LPICP}_i < (1-\alpha)
\right\},
$
 and $\eta$ is the positive user-defined parameter. Finally, the overall average local coverage width-based criterion is computed as
\begin{equation*}
\mathbf{LCWC}
=
\frac{1}{k}
\sum_{i=t+1}^{t+k}
\mathbf{LCWC}_i.
\end{equation*}

\end{enumerate}

\section{ Forecasting Frameworks: An Analysis through the Lens of Calibration Requirements }
In this section, we analyze existing forecasting frameworks through the lens of different levels of calibration granularity.

\subsection{QR based models do not meet GCR}

Before asking whether a principled procedure can enforce
the LCR, a more fundamental question must be addressed: do
existing QR-based models even provide principled guarantees
for the GCR? We show that the answer is negative in the time-
series setting, which further motivates the need for a post-hoc
recalibration framework.

The QR-based probabilistic forecasting models minimize a
pair of pinball loss problems with the quantile parameters $\frac{\alpha}{2}$
and $1 - \frac{\alpha}{2}$, respectively. The pinball loss estimator is a widely
adopted estimator with strong theoretical guarantees, including
asymptotic convergence and consistency properties \citep{takeuchi2006nonparametric} . Given
training set $\{(x_i, y_i) : x_i \in \mathbb{R}^d, y_i \in \mathbb{R}, i = 1, 2, \ldots, n\}$,
sampled from the joint distribution of the random variables
$(X, Y)$, the pinball estimator for $\tau$-th quantile is given by
\begin{equation}
    \hat{f}_\tau(x) := \min_{f \in \mathcal{F}} \sum_{i=1}^n \rho_\tau(y_i - f(x_i)) 
    \label{eq:19}
\end{equation}
that satisfies the asymptotic marginal convergence guarantees
given by
\begin{equation}
    P(Y \leq f_\tau(X)) = \tau, \quad \text{as } n \to \infty, 
    \label{eq:191}
\end{equation}
under two general conditions:
\begin{enumerate}
    \item[(i)] $(x_i, y_i)$ are independently and identically sampled from $(X, Y)$.
    \item[(ii)] The class of functions $\mathcal{F}$ from where the $\hat{f}_\tau$ should be selected should be regular.
\end{enumerate}

Coming back to our quantile-based probabilistic forecasting
setting, the estimated forecast PI bound functions $(f_1, f_2)$
correspond to the conditional $\frac{\alpha}{2}$ and $1 - \frac{\alpha}{2}$ quantiles of
$X_{i+1} \mid H_i$, respectively. These functions are learned by
minimizing the corresponding pinball loss functions over the
sample set
$\{(h_i, x_{i+1}) : h_i \in \mathbb{R}^w, x_{i+1} \in \mathbb{R}, i = w, w+1, \ldots, t\}$,
where $h_i$ contains the most recent $w$ observations at the $i$-th time-stamp.
However, the asymptotic marginal coverage guarantee derived from (\ref{eq:191})
\begin{equation*}
    P(f_1(X) \leq Y \leq f_2(X)) = 1 - \alpha, \quad \text{as } t \to \infty, 
\end{equation*}
requires the pairs $(h_i, x_{i+1})$ to be independently and identically
distributed, along with certain regularity conditions upon the class
of functions from where the quantile pair functions $(f_1, f_2)$ need
to be selected. {Unfortunately, both of these conditions are
difficult to satisfy while learning temporal dependencies  with deep learning models, which means QR-based deep probabilistic forecasting models cannot consistently satisfy even
the GCR in the theoretical sense.} This fundamental limitation
directly justifies the need for a post-hoc conformal recalibration
framework in forecasting tasks.

\subsection{ACI meets the GCR}
The conformal framework \citep{vovk2005algorithmic} provides a general methodology
for transforming any PI into a prediction set that
attains the target coverage with finite samples in a global
sense, under very general data exchangeability assumptions.
However, in the case of time-series data, the exchangeability
assumption is violated, as the data-generating process is supposed
to evolve in an unknown fashion. The standard  Adaptive Conformal
Inference (ACI) framework \citep{gibbs2021adaptive} addresses this limitation by providing a very effective solution for on-line data stream.

In ACI, a probabilistic forecast is first obtained from the warm-up samples
$\{x_j\}_{j=1}^{T_{\mathrm{int}}}$ using any base model.
 Thereafter, for the stream of online observations
$\{ x_{T_{int}+\hat{i}}\}_{\hat{i} =1} ^{t}$, ACI lets the miscoverage level
$\alpha_j$ evolve through the following recursion
\begin{equation}
    \alpha_{\hat{i}+1} = \alpha_{\hat{i}} + \gamma\!\left(\alpha - \mathbb{I}\{x_{i+1} \notin \hat{C}^{(\alpha_{\hat{i}})} (h_i)\}\right), 
    \label{aci}
\end{equation}
where $\hat{C}^{(\alpha_{\hat{i}})} (h_i)$ denotes the conformal prediction set estimated for $x_{i+1}$ targeting the coverage level $1-\alpha_{\hat{i}}$, where $i= \hat{i}+T_{int}$. Here, $0 < \gamma < 1$ is a fixed user-defined step size that continuously adjusts the miss-coverage level, enabling the conformal prediction set to adapt to distributional shifts over time. The initial level of miscoverage in recursion (\ref{aci}) is fixed to $\alpha_1= \alpha$. Thereafter, at each step $i$, the conformal correction is 
recomputed using the updated miscoverage level $\alpha_{\hat{i}}$ without retraining the base forecasting model.

An important key theoretical guarantee for standard ACI (\ref{aci}) is noted in Proposition~4.1 of \citep{gibbs2021adaptive}. It states that regardless of the underlying data-generating
process, the following bound holds true
\begin{small}
\begin{equation}
    \left|\frac{1}{t} \sum_{k=T_{int}}^{t} \mathbb{I}\{x_{k+1} \notin \hat{C}^{(\alpha_{\hat{k}})} (h_k)\} - \alpha\right|
    \leq \frac{\max(\alpha_1, 1-\alpha_1) + \gamma}{\gamma t}.
    \label{eq:22}
\end{equation}
\end{small}
The above inequality bounds the miss-coverage error obtained through the standard ACI process (\ref{aci}) with $0$   as $t \xrightarrow{} \infty$, in view of the fact that the $\max(\alpha_1, 1-\alpha_1)$ and $\gamma$ is bounded above with $1$.  The ACI prediction set attains the GCR under
arbitrary distributional shift, without relying on the i.i.d.\ or
regularity conditions. It makes the ACI estimator a natural post-hoc remedy for 
ensuring the GCR in forecasting models in theoretical sense.

To smooth the  $\alpha_{\hat{i}}$ trajectories, \cite{gibbs2021adaptive} further suggest the recency-weighted variant of ACI that  updates the $\alpha_{\hat{i}}$ using a weighted aggregation of past coverage errors, with greater weight assigned to more recent errors using
\begin{equation}
    \alpha_{\hat{i}+1}
    =
    \alpha_{\hat{i}}
    +
    \gamma
    \left(
        \alpha
        -
        \sum_{\hat{s}=1}^{i} w_{\hat{s}} \, \mathrm{err}_{\hat{s}}
    \right),
    \label{waci}
\end{equation}
where $\mathrm{err}_{\hat{s}} = \mathbb{I}\{x_{s+1} \notin \hat{C}^{(\alpha_{\hat{s}})} (h_s)\}$ denotes the coverage error at step ${\hat{s}}$, and $\{w_{\hat{s}}\}_{1 \leq {\hat{s}} \leq \hat{i}} \subseteq [0,1]$ is a sequence of increasing weights satisfying $\sum_{\hat{s}=1}^{\hat{i}} w_{\hat{s}} = 1.$
In particular, we take the weights to decay exponentially with the time elapsed since step $\hat{s}$, given by
    $w_{\hat{s}}=
    \frac{M^{\,i-\hat{s}}}
    {\displaystyle \sum_{s'=1}^{i} M^{\,i-s'}}$, where $M$ is chosen to be $0.95$ in \citep{gibbs2021adaptive}.

\citet{gibbs2021adaptive} observe that, although the weighted update~(\ref{waci})
produces a smoother trajectory of $\alpha_{\hat{i}}$, it does not yield noticeable
performance gains in their experiments. Furthermore, whether the weighted ACI process~(\ref{waci}) satisfies the GCR has not been answered  in the ACI literature.  We close this gap in Proposition 3 of Appendix A, which shows that the weighted ACI process~(\ref{waci}) attains the GCR under arbitrary distribution shift.

\subsection{ACI and LCR}
Now, we analyze the ACI framework from the lens of the LCR.
 Recall from Section~III that the forecast PI $[f_1,f_2]$
is said to be locally well-calibrated at time-index $i$ over a window of size $w$ if
\[
    P\bigl(f_1(H_i) \le X_{i+1} \le f_2(H_i) \mid H_i^{(w)}\bigr) \;\ge\; 1-\alpha,
\]
i.e., the empirical coverage restricted to the neighborhood $\mathcal{N}_w(h_i)$ must itself meet the
nominal target. 

We note that the ACI recursion (\ref{aci})
satisfies the GCR asymptotically through (\ref{eq:22}) under arbitrary, unknown distribution shift, without relying on
exchangeability . However, it says
nothing about miscoverage realized within any particular local window that is actually required in most of the energy and other relevant applications for reliable decision support. Proposition~\ref{prepo1}
below connects this gap by evaluating the target of the ACI recursion in view of the local calibration ability.  

\begin{proposition}
\label{prop:lpicp_identity} 
Let us consider the standard ACI process (\ref{aci}) operated upon the time-series observations $\{ x_{T_{int}+\hat{i}}\}_{\hat{i} =1} ^{t}$ streamed online
, then irrespective of the data-generating process,
the local empirical coverage produced by the ACI recursion satisfies 
\begin{equation}
    (1 - \alpha) - \mathrm{LPICP}^{(L)}_{\hat{i}} 
    = -\frac{\Delta\alpha_{\hat{i}}}{\gamma w}, 
    \qquad 
   \mbox{where~~~~} \Delta\alpha_{\hat{i}} := \alpha_{\hat{i} +1} - \alpha_{\hat{i} -w+1},
    \label{eq:lpicp_identity_prop}
\end{equation}
and consequently
\begin{equation}
    \mathrm{LE}^{(L)}_{\hat{i}} 
    = \max\!\left(-\frac{\Delta\alpha_{\hat{i}}}{\gamma w},\; 0\right).
    \label{eq:le_identity_prop}
\end{equation}


\label{prepo1}
\end{proposition}

For proof, please see Appendix A.

\textbf{Interpretations of Proposition \ref{prepo1} :-} For the given $\hat{i}^{th}$ local window, Proposition \ref{prepo1} connects the amount of local miscoverage error (LE$_{\hat{i}}$) with the ratio of negative drift $(-\Delta \alpha)$ and $\gamma w$. If most of the $err_t$ computed across the $\hat{i}^{th}$ local window remains zero in the ACI recursion, the $\hat{i}^{th}$ local window is said to be over-covered. In such cases, $\Delta \alpha$ turns out to be a positive number, making the ACI estimate well-calibrated for the $\hat{i}^{th}$ local window. However, in the case of under-coverage across the $\hat{i}^{th}$ local window, the local error $LE_{\hat{i}}$ is determined through $\frac{-\Delta \alpha}{\gamma w}$, which reduces linearly with the size of the local window $w$.

\textbf{Impact of $\gamma$ upon the error:-}. The user-defined parameter $\gamma$ in the ACI recursion should be chosen to appropriately track the rate of distributional drift in order to obtain well-calibrated intervals. Increasing $\gamma$ leads to more aggressive adaptation of the interval over time, which can improve global calibration (as evident from Eq.~\ref{eq:22}), but at the cost of increased interval width. Recent variants of ACI, such as AgACI \citep{pmlr-v162-zaffran22a} and DtACI\citep{gibbs2024conformal}, address this issue by adaptively tuning $\gamma$ to improve calibration. However, optimizing $\gamma$ primarily controls the overall adaptation rate and does not directly target local calibration errors $LE_{\hat{i}}$. In particular, as indicated by Proposition~\ref{prepo1}, increasing $\gamma$ can also increase the accumulated change $\Delta\alpha$, potentially widening the intervals without effectively reducing local calibration error.

\section{ Locally Adaptive Conformal Inference (LACI) for Probabilistic Forecasting}
\label{sec5}
In view of the limitations of ACI and its existing variants in controlling local calibration errors, we realize the need to improve the ACI process. We first require the identification of the condition where the ACI processes often leads to the local coverage failure.

Although the    ACI and its variants can effectively accommodate the gradual distributional shifts over time,  our observation entails that the abrupt changes in the autoregressive distribution of a time-series signal may lead to abrupt local coverage failure.

To formalize this behavior clearly, consider an ACI-generated
coverage-error trajectory
$\{\mathrm{err}_{\hat{s}}\}_{{\hat{s}}=1}^{\hat{i}}$, where
$\mathrm{err}_{\hat{s}} \in \{0,1\}$.
Let
\[
\mathcal{F}_{\hat{i}} =
\{ \hat{s} \in \{1,\ldots,\hat{i}\} : \mathrm{err}_{\hat{s}} = 1\}
\]
denote the set of historical coverage-failure indices up to
time $\hat{i}$. 
Further, for each historical index $\hat{s}$, let $\mathrm{LE}_{\hat{s}}^{(L)}$
denote the corresponding left-centered local coverage error.
We define the recency-weighted average local error up to
time $\hat{i}$ as
\begin{equation}
\overline{\mathrm{LE}}_{\hat{i}}^{(L)}
=
\frac{
\sum_{k=1}^{\hat{i}} M^{\hat{i}-k}\mathrm{LE}_k^{(L)}
}{
\sum_{k=1}^{\hat{i}} M^{\hat{i}-k}
},
\label{eq:weighted_local_error}
\end{equation}
where $M=0.95$ is the one good choice of exponential decay factor.

We characterize the abrupt changes in the auto-regressive distribution through the abrupt local coverage errors. While $\hat{i}^{th}$ ACI update, we identify the abrupt  local coverage errors at index $\hat{s}$, $1 \leq \hat{s} \leq \hat{i}$, with condition $\mathrm{LE}_{\hat{s}}^{(L)} > \overline{\mathrm{LE}}_{\hat{i}}^{(L)}$. Further, we claim that local coverage error trajectory of the ACI process satisfy the following condition
\begin{equation}
\sum_{\hat{s}\in\mathcal{F}_{\hat{i}}}
\mathbb{I}
\left\{
\mathrm{LE}_{\hat{s}}^{(L)}
>
\overline{\mathrm{LE}}_{\hat{i}}^{(L)}
\right\}
\geq 
\frac{|\mathcal{F}_{\hat{i}}|}{p},
\label{ass:severity-dominance}
\end{equation}
where $|\mathcal{F}_{\hat{i}}|$ is the cardinality of the set $\mathcal{F}_{\hat{i}}$ and $p \geq 1$. In other words, in ACI recursion, a significant majority of
historical coverage-failure points cause the abrupt local coverage error in the sense that their local coverage errors are greater than the weighted average of local errors. We empirically show that the above inequality (\ref{ass:severity-dominance}) remains valid with at least $p < 2$ in Section \ref{sec:empirical}.



Further, we note that the condition (\ref{ass:severity-dominance}) holds  both for the  standard ACI and Weighted ACI process. A natural question arises that can we develop the weighting scheme  for coverage errors in ACI process beyond the recency for possible improvement in the overall local calibration ?

Our Local Adaptive Conformal Inference (LACI) process addresses this question well by assigning the higher weights to  coverage failure points satisfying (\ref{ass:severity-dominance}).  The LACI process modifies the ACI update to improve local calibration as follows

   \begin{equation}
    \alpha_{{\hat{i}}+1}
    =
    \alpha_{\hat{i}}
    +
    \gamma
    \left(
        \alpha
        -
        \sum_{\hat{s}=1}^{{\hat{i}}}\widetilde{w}_s \, \mathrm{err}_{\hat{s}}
    \right),
    \label{wlaci}
\end{equation} 
where 
\begin{equation}
    \widetilde{w}_{s}
=
\frac{0.95^{\hat{i}-{\hat{s}}} \left(LE_{\hat{s}}^{(L)}\right)}{\displaystyle\sum_{k=1}^{\hat{i}}0.95^{\hat{i}-k}\left(LE_k^{(L)}\right)},
\qquad 1\leq \hat{s} \leq \hat{i} .
\end{equation}





\subsection{The LACI Algorithim}
We present the detail implementation of the LACI algorithm as follows.

\begin{algorithm}[H]
\caption{LACI: Locally Adaptive Conformal Inference}
\label{alg:laci}
\begin{algorithmic}[1]

\State \textbf{Data:}  Warm-up batch time-series observation$\{x_i\}_{i=1}^{T_{int}}$, stream of online observations
$\{ x_{T_{int}+\hat{i}}\}_{\hat{i} =1} ^{t}$,
significance level $\alpha$,
adaptation parameter $\gamma$, decay factor $M=0.95$,
local window size $w$,  warm-up sample size
$T_{\mathrm{init}}\geq w$ and $\alpha_1 = \alpha$.

\For{$\hat{i}=1,\ldots,t$}
  \State    ${i} = \hat{i}+ T_{int}$
    \State Construct the conformal set
    $\hat{C}^{(\alpha_{\hat{i}})}(h_{i})$
    targeting $x_{{{i}}+1}$ with coverage
    $1-\alpha_{\hat{i}}$.

    \State $\displaystyle
    err_{\hat{i}}\gets
    \mathbb{I}\left\{
    x_{i+1}
    \notin
    \hat{C}^{(\alpha_{\hat{i}})}(h_{i})
    \right\}$.

    \State $\displaystyle
    \mathrm{LPICP}^{(L)}_{\hat{i}}\gets
    \frac{1}{w}\sum_{j=\hat{i}-w+1}^{\hat{i}}
    \mathbb{I}\left\{
    x_{T_{\mathrm{init}}+j+1}
    \in
    \hat{C}^{(\alpha_{\hat{i}})}(h_{T_{\mathrm{init}}+j})
    \right\}$.

    \State $\displaystyle
    LE_{\hat{i}}^{(L)}\gets
    \max\left\{
    (1-\alpha)-\mathrm{LPICP}^{(L)}_{\hat{i}},0
    \right\}$.

    \For{$\hat{s}=1,\ldots,\hat{i}$}
        \State $\displaystyle
        \widetilde{w}_{\hat{s}}\gets
        \frac{M^{\hat{i}-\hat{s}}LE_{\hat{s}}^{(L)}}
        {\sum_{k=1}^{\hat{i}}M^{{\hat{i}}-k}LE_k^{(L)}}$.
    \EndFor

    \State $\displaystyle
    \widehat{e}_{\hat{i}}\gets
    \sum_{\hat{s}=1}^{\hat{i}}\widetilde{w}_{\hat{s}}\,err_{\hat{s}}$.

    \State $\displaystyle
    \alpha_{\hat{i}+1}\gets
    \alpha_{\hat{i}}+\gamma(\alpha-\widehat{e}_{\hat{i}})$.

\EndFor

\end{algorithmic}
\end{algorithm}

\subsection{Analysis of the LACI Algorithm}
The central idea of LACI is that, when miscoverage occurs, the change in
$\alpha_{\hat{i}}$ should reflect not only whether past errors occurred, but also how
severe the local coverage failure was at the time. Standard and weighted ACI
treat every miss alike, as a binary error $\mathrm{err}_{\hat{s}}\in\{0,1\}$, weighted
at most by its recency. LACI additionally weights each error by the local
error $\mathrm{LE}^{(L)}_{\hat{s}}$ observed at that time, so that misses occurring
during severe local failures dominate the update of $\alpha_{\hat{i}}$. In other words, LACI assign disproportionately larger weights to more severe local errors while computing the $\alpha$ trajectories. It enables the LACI to improve the local calibration ability of the existing ACI model whenever the latter suffers the condition \ref{ass:severity-dominance}. The following lemma formalizes LACI behavior.

\begin{lemma}
Consider an ACI-generated fixed binary coverage-error trajectory  $\{\mathrm{err}_{\hat{s}}\}_{{\hat{s}}=1}^{\hat{i}}$, $\mathrm{err}_{\hat{s}} \in\{0,1\}$ with the local coverage error $LE_{\hat{s}}^{(L)}$ at time $\hat{s}$ and  recency-weighted
average of the local coverage errors till the $\hat{i}^{th}$ index $\overline{LE}_{\hat{i}}^{(L)}$ of (\ref{eq:weighted_local_error}).
Also $w_{\hat{s}}$ and $\widetilde{w}_{\hat{s}}$ are the weights in  ACI process (\ref{waci})  and LACI process (\ref{wlaci}), respectively. Then
\begin{equation}
LE_{\hat{s}}^{(L)} > \overline{LE}_{\hat{i}}^{(L)}
\Longrightarrow\quad
\widetilde{w}_{\hat{s}} > w_{\hat{s}} \mbox{,} ~1 \leq {\hat{s}} \leq \hat{i}.
\label{lemma1}    
\end{equation}

\end{lemma}


The proof of the above lemma is provided in Appendix A.

Further, under the LACI algorithm, the assignment of a larger weight in the presence of a sudden drift in the time-series signal, characterized by the condition (\ref{ass:severity-dominance}), enables improved local calibration ability . This is formally stated in the following proposition.


\begin{proposition}
\label{prop:selective-amplification}
If $\{\mathrm{err}_{\hat{s}}\}_{s=1}^{\hat{i}}$ is the error trajectory obtained by the ACI process till $\hat{i}^{th}$ time index satisfying the condition (\ref{ass:severity-dominance}), then use of the LACI recursion with the same initial miscoverage $\alpha_1$, guarantees the improvement in the local calibration at the subsequent $(\hat{i}+1)^{th}$ time index as $p \rightarrow 1$.

\end{proposition}

The proof of the above proposition is provided in Appendix A.

\section{Empirical Analysis}
\label{sec:empirical}

In this section, we present a comprehensive empirical evaluation that systematically verifies the claims made throughout the paper. Before presenting the numerical results, we detail our experimental setup as follows.


\subsection{Experimental Setup and Datasets}
\label{subsec:multidomain_setup}




\label{subsec:multidomain_setup}

All datasets are evaluated under a unified sliding-window online
forecasting framework to ensure a consistent and comparable
assessment across domains. Let $\{x_i\}_{i=1}^{N}$ denote a univariate time series containing $N$
observations. The first $T_{\mathrm{int}}$ observations are reserved for the
initial model development stage, while the subsequent observations
$\{x_{T_{\mathrm{int}}+\hat{i}}\}_{\hat{i}=1}^{t}$, with $t = N - T_{\mathrm{int}}$,
are processed sequentially in an online forecasting setting with the initial coverage $\alpha_1$. At each online
step $\hat{i}$, the model generates a interval for
$x_{T_{\mathrm{int}}+\hat{i}}$ using only the information available up to time
$T_{\mathrm{int}}+\hat{i}-1$, thereby preventing information leakage from
future observations.

The underlying predictive distribution is approximated using
quantile regression with deep learning architectures, including
Long Short-Term Memory (LSTM), Gated Recurrent Unit (GRU), and
Temporal Convolutional Network (TCN). For each architecture,
two separate forecasting models are trained to estimate the
lower and upper conditional quantiles corresponding to a nominal
coverage level of $1-\alpha=0.90$. Specifically, the models
estimate the $\alpha/2$ and $1-\alpha/2$ quantiles by minimizing
the pinball loss. Given an input sequence $x_i$, the resulting
quantile estimates are denoted by
$\hat{q}_{\alpha/2}(x_i)$ and
$\hat{q}_{1-\alpha/2}(x_i)$, respectively.


At each online prediction step, the empirical quantile of the
non-conformity scores given in (\ref{eq:conformity_score}) of \cite{romano2019conformalized} is computed at the adaptively determined confidence level. This conformal correction is subsequently
applied to the lower and upper quantile functions of  predicted PI to obtain
the calibrated PI. The adaptive methods
considered in this study, including Adaptive Conformal
Inference (ACI) and its variants, update the miscoverage level
over time to respond to changes in the underlying data
distribution.

Further, all of the used forecasting
models were periodically fine-tuned using newly available
observations at six-step intervals (corresponding to six hours
in the hourly evaluation setting). Between successive updates,
the model parameters remain fixed, while the conformal
calibration and interval adaptation procedures continue to
operate at every prediction step. This separation allows the
forecasting model to capture evolving temporal dependencies
while the conformal adaptation mechanisms respond to changes
in prediction errors and empirical coverage.

All experiments target a nominal coverage probability of
$1-\alpha=0.90$. In addition to global coverage and interval-width
measures, local coverage behaviour is evaluated using a rolling
window of $w=24$ observations, corresponding to a 24-hour period
for hourly data.   Codes and datasets used in this study are publicly
available on anonymous GitHub:\url{https://github.com/Manush-2005/TLMR_Local_Gitub}.









\begin{table}[H]
\centering
{\small
\renewcommand{\arraystretch}{1.2}
\begin{tabular}{llccc}
\hline
\textbf{Type} & \textbf{Dataset} & \textbf{No. of Points} & 
\textbf{Time Frame} & \textbf{Location} \\
\hline

\multirow{6}{*}{Wind Power}
& 50Hertz\citep{jorge_gustavo_sandoval_sim_o_netztransparenz_2021} 
& 9,957 & 1 year  & Germany \\
& TransnetBW\citep{jorge_gustavo_sandoval_sim_o_netztransparenz_2021} 
& 9,957 & 1 year  & Germany \\
& TenneT TSO\citep{jorge_gustavo_sandoval_sim_o_netztransparenz_2021} 
& 9,957 & 1 year  & Germany \\
& Amprion\citep{jorge_gustavo_sandoval_sim_o_netztransparenz_2021} 
& 9,957 & 1 year  & Germany \\
& Location 1\citep{rahim2017wind} 
& 9,929 & 2 years & Beijing, China \\
& Location 2\citep{rahim2017wind} 
& 9,929 & 2 years & Beijing, China \\

\hline

Hydro Power
& HydroPower\citep{al_shahriar_hossain_2026} (Granularity: Daily)
& 1,782 & 4 years & Bangladesh \\

\hline

\multirow{2}{*}{Solar Power}
& Solar Plant 1\citep{Ani2020Solar}
& 9,308 & 34 days & India \\
& Solar Plant 2\citep{Ani2020Solar}
& 9,561 & 34 days & India \\

\hline
\end{tabular}}
\caption{Summary of the datasets used in this study.}
\label{dataset_summary}
\end{table}

\textbf{Datasets:-} In our experiments, we consider popular forecasting challenges that arise in various renewable power generation domains such as wind, solar and hydro power. All of our datasets are univariate and contain the hourly recording of energy variables.
A detailed summary of the datasets, including the number of observations, temporal coverage,  and geographical location, is provided in Table~\ref{dataset_summary}.

\subsection{Understanding LACI Behaviour under Abrupt Distribution Shift}
Before presenting results on the benchmark energy datasets under the
experimental setup described above, we conduct a simple synthetic experiment
based on an AR(1) process to illustrate the behaviour of LACI relative to ACI.

\begin{figure}[!ht]
    \centering

    \includegraphics[width=0.92\textwidth]
    {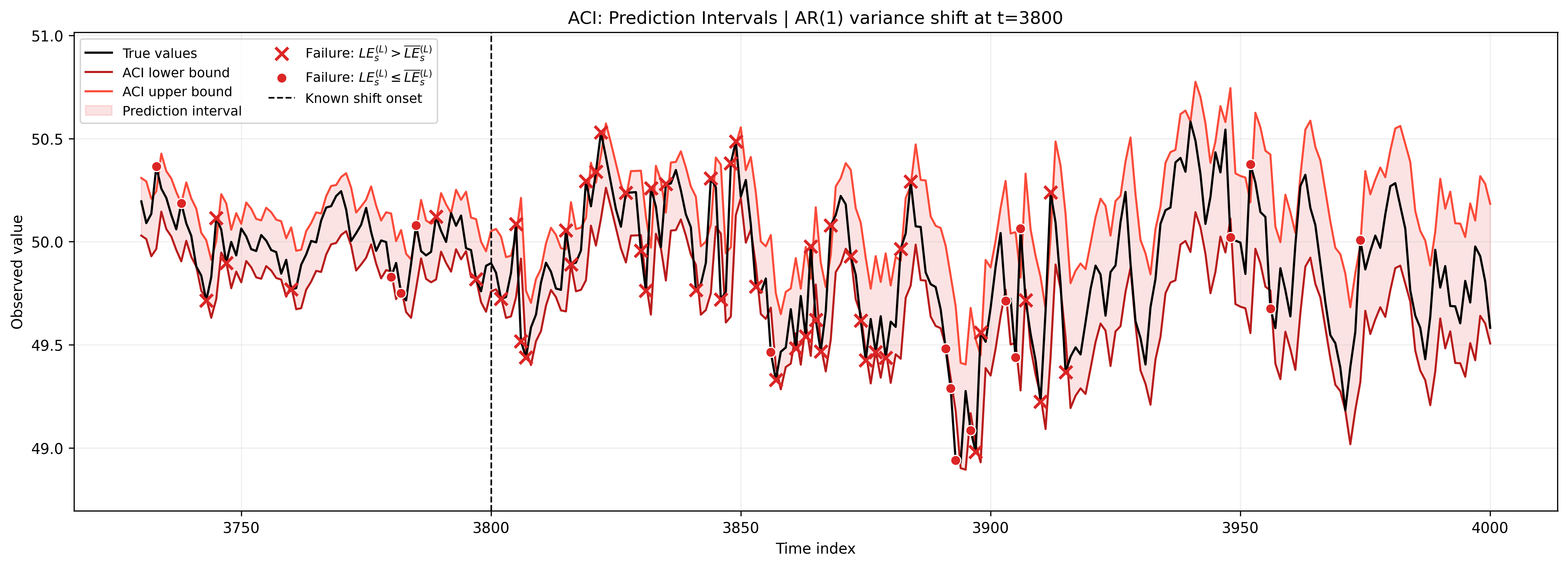}

    \vspace{0.2cm}

     \includegraphics[width=0.92\textwidth]
    {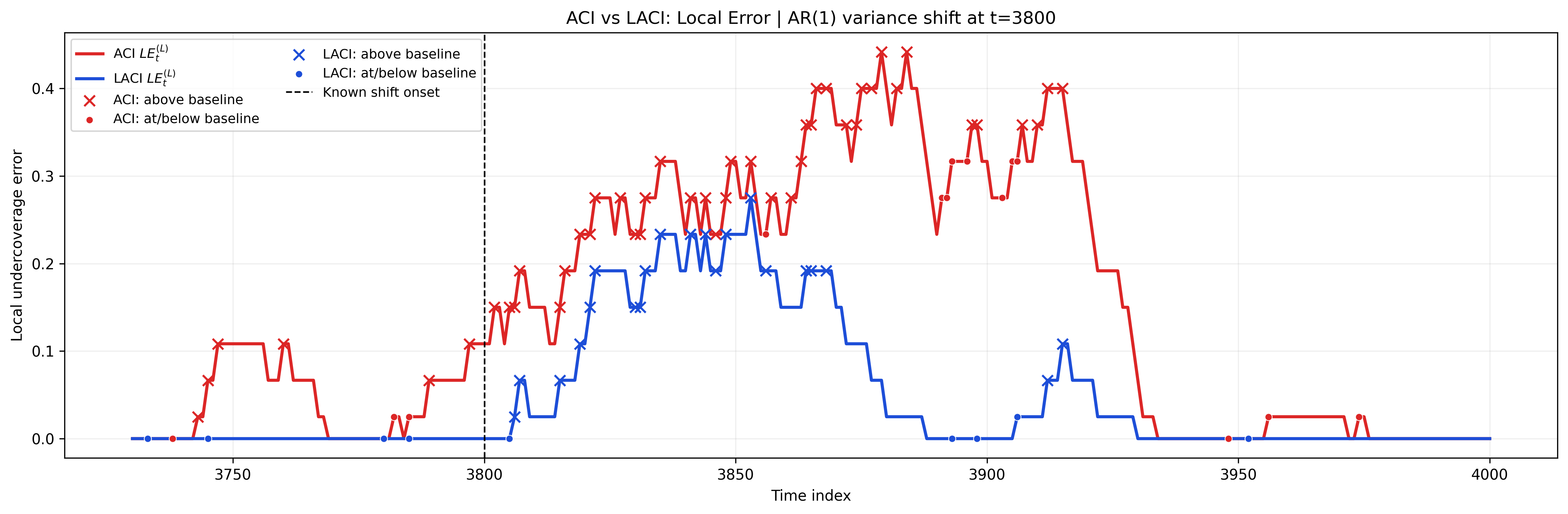}
    
       \vspace{0.2cm}

    \includegraphics[width=0.92\textwidth]
    {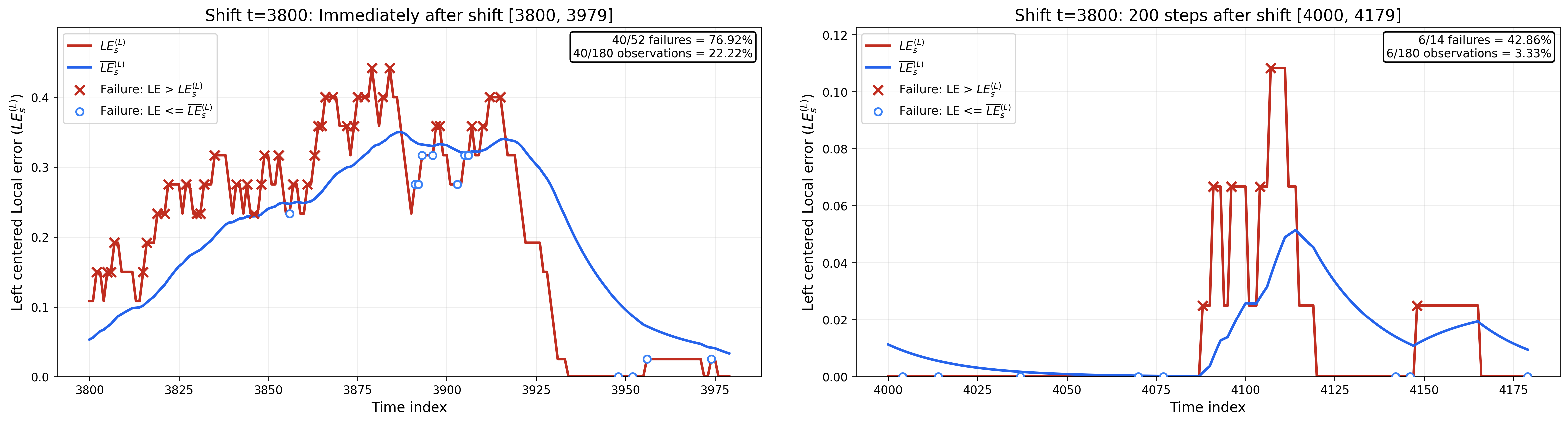}

    \caption{Comparison of ACI and LACI under an abrupt
    variance shift in a synthetic AR(1) time series at
    $t=5800$. (Top) Most of the local error coverage obtained by the ACI process is abrupt i.e.\
$\mathrm{LE}_s^{(L)} > \overline{\mathrm{LE}}_s^{(L)}.$
    (Middle) Local coverage error trajectory of ACI and LACI mehtod, LACI tends to improve the local calibration in case of the abrupt changes in the auto-regressive distribution. (Bottom) Abrupt coverage errors
    are observed to be significantly higher in the interval (of size 180) immediately after the abrupt distribution shift over the normal case where there is no such abrupt distribution shift.}

    \label{fig:synthetic_shift}
\end{figure}

We generate the 7200 observations with the AR(1) like  process with abrupt changes in noise distribution using
\begin{equation*}
    x_t = c + \phi x_{t-1} + \epsilon_t,
    \qquad
    \epsilon_t \sim \mathcal{N}(0,\sigma_t^2),
\end{equation*}
where $c=10$, $\phi=0.8$, and $x_0=50$. 
To introduce the abrupt distribution shift, we introduce the five variance shifts  at
$t=\{1800,2800,3800,4800,5800\}$ i,e 
we pick the $\sigma$ of the noise distribution at these points from set
$
\{0.06,\; 0.17,\; 0.09,\; 0.21,\; 0.10,\; 0.23\}$ respectively.

At first, the AR(1) point forecasting model is fitted using ordinary least squares on the first 700 observations to learn the value of parameters c and $\phi$ . The fitted model is then used to generate one-step-ahead point forecasts, while ACI and LACI
independently and adaptively construct conformal intervals using a rolling
calibration window of 300 observed residual scores. Both methods
use the same adaptation parameter $\gamma=0.001$, exponential
decay factor $M=0.95$, target coverage $1-\alpha=0.90$, and
local window size $w=24$. 

\begin{table*}[htbp]
\centering
\small
\begin{tabular}{cccc}
\hline
\textbf{Window} &
\textbf{Coverage Failures } &
\textbf{\#($\mathrm{LE}_{\hat{s}}^{(L)} > \overline{\mathrm{LE}}_{\hat{s}}^{(L)}$)} &
\textbf{(\%) of \textbf{\#($\mathrm{LE}_{\hat{s}}^{(L)} > \overline{\mathrm{LE}}_{\hat{s}}^{(L)}$)}} \\
\hline
1800--1979 & 54 & 36 & 66.67 \\
2000--2179 & 11 & 1 & 9.09 \\
\hline
3800--3979 & 52 & 40 & 76.92 \\
4000--4179 & 14 & 6 & 42.86 \\
\hline
5800--5979 & 50 & 37 & 74.00 \\
6000--6179 & 16 & 7 & 43.75 \\
\hline
\end{tabular}
\caption{Local coverage errors of ACI under synthetic variance shifts. In the
window immediately after a shift in the noise variance, most local coverage
errors are abrupt, i.e.\
$\mathrm{LE}_s^{(L)} > \overline{\mathrm{LE}}_s^{(L)}$, compared with a typical
window of the same length. $\#$ denotes the number of time steps satisfying
the stated condition.}
\label{legeqls}
\end{table*}

\begin{table}[htbp]
\centering

\label{tab:local_coverage_comparison}
\begin{tabular}{c|cc|cc|cc}
\hline
\multirow{2}{*}{Shift Time} 
& \multicolumn{2}{c|}{LPICP} 
& \multicolumn{2}{c|}{Mean LE} 
& \multicolumn{2}{c}{LMPIW} \\
\cline{2-7}
& ACI & LACI & ACI & LACI & ACI & LACI \\
\hline
1800 & 0.7000 & 0.8556 & 0.19875 & 0.07097 & 0.37307 & 0.50873 \\
3800 & 0.7111 & 0.8556 & 0.20537 & 0.07097 & 0.48624 & 0.66982 \\
5800 & 0.7222 & 0.8556 & 0.18421 & 0.05958 & 0.55724 & 0.69948 \\
\hline
\end{tabular}

\caption{Local performance comparison of ACI and LACI 
during selected undercoverage periods.}
\label{tab2}
\end{table}

At first, we observe the local coverage error trajectory after each distribution shift over the first 180 observations
and compute the fraction of the observations that satisfy abrupt increase
in its local coverage error characterized through $\mathrm{LE}_s^{(L)} >
\overline{\mathrm{LE}}_s^{(L)}$. This fraction is significantly greater than the fraction of the observations satisfying $\mathrm{LE}_s^{(L)} >
\overline{\mathrm{LE}}_s^{(L)}$ in the window that does not suffer such distribution shift. Table~\ref{legeqls} provides empirical evidence for this behavior. It
compares how often the local coverage error of ACI increases abruptly, i.e.\
$\mathrm{LE}_s^{(L)} > \overline{\mathrm{LE}}_s^{(L)}$, in the window of length
180 immediately after the variance shift with how often it does so in a
typical window of the same length. After the variance shift, the abrupt coverage errors are observed to be significantly higher than normal case that is also well illustrated in Figure \ref{fig:synthetic_shift} (c) for a particular case.

We further compare LACI with ACI over the first 180 observations after each
variance shift. Table~\ref{tab2} reports their local performance and shows
that LACI substantially improves on ACI in this regime.
Figure~\ref{fig:synthetic_shift}(a) shows that, after the variance shift in
the AR process at $t=3800$, most of the local coverage errors of ACI are
abrupt, i.e.\ $\mathrm{LE}_s^{(L)} > \overline{\mathrm{LE}}_s^{(L)}$.
Figure~\ref{fig:synthetic_shift}(b) shows the local error trajectories of ACI
and LACI after the shift, and LACI markedly reduces the magnitude of the
local errors.

\subsection{ Adaptive Conformal Inference provides better local calibration than Quantile Regression  methods.}
\label{subsubsec:claim1}

In this subsection, we show that Adaptive Conformal Inference (ACI) outperforms the QR-based methods in terms of our local calibration metrics. Under distribution shifts commonly encountered in energy forecasting, ACI can adapt to changing data distributions through its dynamic update of the parameter $\alpha$, enabling it to maintain better local calibration over time.

\begin{table}[H]
\centering
\vspace{-2mm}
{\fontsize{8pt}{9.5pt}\selectfont
\setlength{\tabcolsep}{4pt}
\renewcommand{\arraystretch}{1.0}
\begin{tabular}{ll|ccc|cc}
\toprule
\textbf{Dataset} & \textbf{Method} &
\multicolumn{3}{c|}{\textbf{Local Metrics}} &
\multicolumn{2}{c}{\textbf{Global Metrics}} \\
\cmidrule(lr){3-5} \cmidrule(lr){6-7}
& &
\textbf{LPIA} & \textbf{LPICE} & \textbf{LMPIW} &
\textbf{PICP} & \textbf{MPIW} \\
\midrule

\multirow{2}{*}{50Hertz}
& Normal\_QR & 0.16 & 0.16 & \textbf{31.46} & 0.75 & \textbf{31.48} \\
& ACI       & \textbf{0.67} & \textbf{0.04} & 49.10 & \textbf{0.90} & 49.12 \\

\midrule

\multirow{2}{*}{TransnetBW}
& Normal\_QR & \textbf{0.71} & \textbf{0.02} & 12.23 & \textbf{0.92} & 12.22 \\
& ACI       & 0.62 & 0.03 & \textbf{11.82} & 0.90 & \textbf{11.81} \\

\midrule

\multirow{2}{*}{TenneTTSO}
& Normal\_QR & 0.22 & 0.12 & \textbf{28.87} & 0.79 & \textbf{28.85} \\
& ACI       & \textbf{0.62} & \textbf{0.04} & 39.99 & \textbf{0.90} & 39.99 \\

\midrule

\multirow{2}{*}{Amprion}
& Normal\_QR & 0.30 & 0.12 & \textbf{21.78} & 0.79 & \textbf{21.79} \\
& ACI       & \textbf{0.68} & \textbf{0.04} & 32.03 & \textbf{0.91} & 32.05 \\

\midrule

\multirow{2}{*}{Location 1}
& Normal\_QR & \textbf{0.66} & 0.04 & 0.03 & 0.89 & 0.03 \\
& ACI       & 0.62 & \textbf{0.04} & \textbf{0.03} & \textbf{0.90} & \textbf{0.03} \\

\midrule

\multirow{2}{*}{Location 2}
& Normal\_QR & \textbf{0.74} & \textbf{0.03} & 0.03 & 0.92 & 0.03 \\
& ACI       & 0.65 & 0.03 & \textbf{0.03} & \textbf{0.90} & \textbf{0.03} \\

\midrule

\multirow{2}{*}{Hydropower}
& Normal\_QR & 0.35 & 0.30 & \textbf{116.75} & 0.62 & \textbf{115.66} \\
& ACI       & \textbf{0.64} & \textbf{0.06} & 184.77 & \textbf{0.88} & 184.92 \\

\midrule

\multirow{2}{*}{Solar Plant 1}
& Normal\_QR & 0.24 & 0.33 & \textbf{225.69} & 0.59 & \textbf{225.72} \\
& ACI       & \textbf{0.62} & \textbf{0.07} & 1442.91 & \textbf{0.88} & 1442.47 \\

\midrule

\multirow{2}{*}{Solar Plant 2}
& Normal\_QR & 0.11 & 0.47 & \textbf{1451.34} & 0.43 & \textbf{1454.35} \\
& ACI       & \textbf{0.48} & \textbf{0.05} & 5350.74 & \textbf{0.88} & 5359.85 \\

\bottomrule
\end{tabular}
}
\vspace{-2mm}
\caption{Comparison of local calibration and global calibration metrics of Normal\_QR and ACI across nine energy datasets using an LSTM architecture with target coverage $1-\alpha=0.90$. Best results are highlighted in bold.}
\label{tab:qr_aci_local_global}
\end{table}

\begin{figure}[H]
    \centering
    \begin{subfigure}[b]{0.50\textwidth}
        \centering
        \includegraphics[width=\textwidth ,height = 0.8\textwidth ]{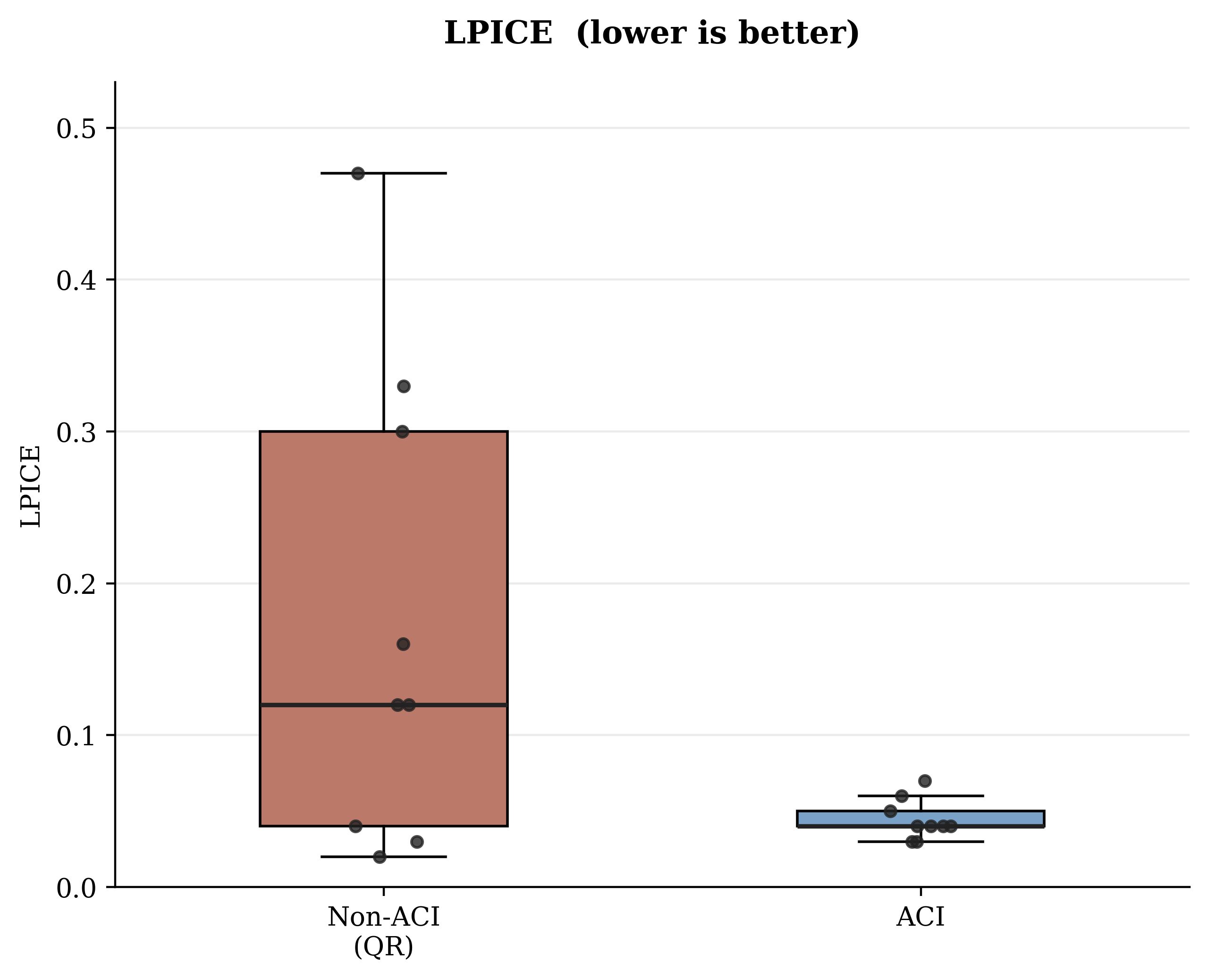}
        \caption{Distribution}
        \label{fig:boxplot}
    \end{subfigure}\hfill
    \begin{subfigure}[b]{0.50\textwidth}
        \centering
        \includegraphics[width=\textwidth, height= 0.8\textwidth]{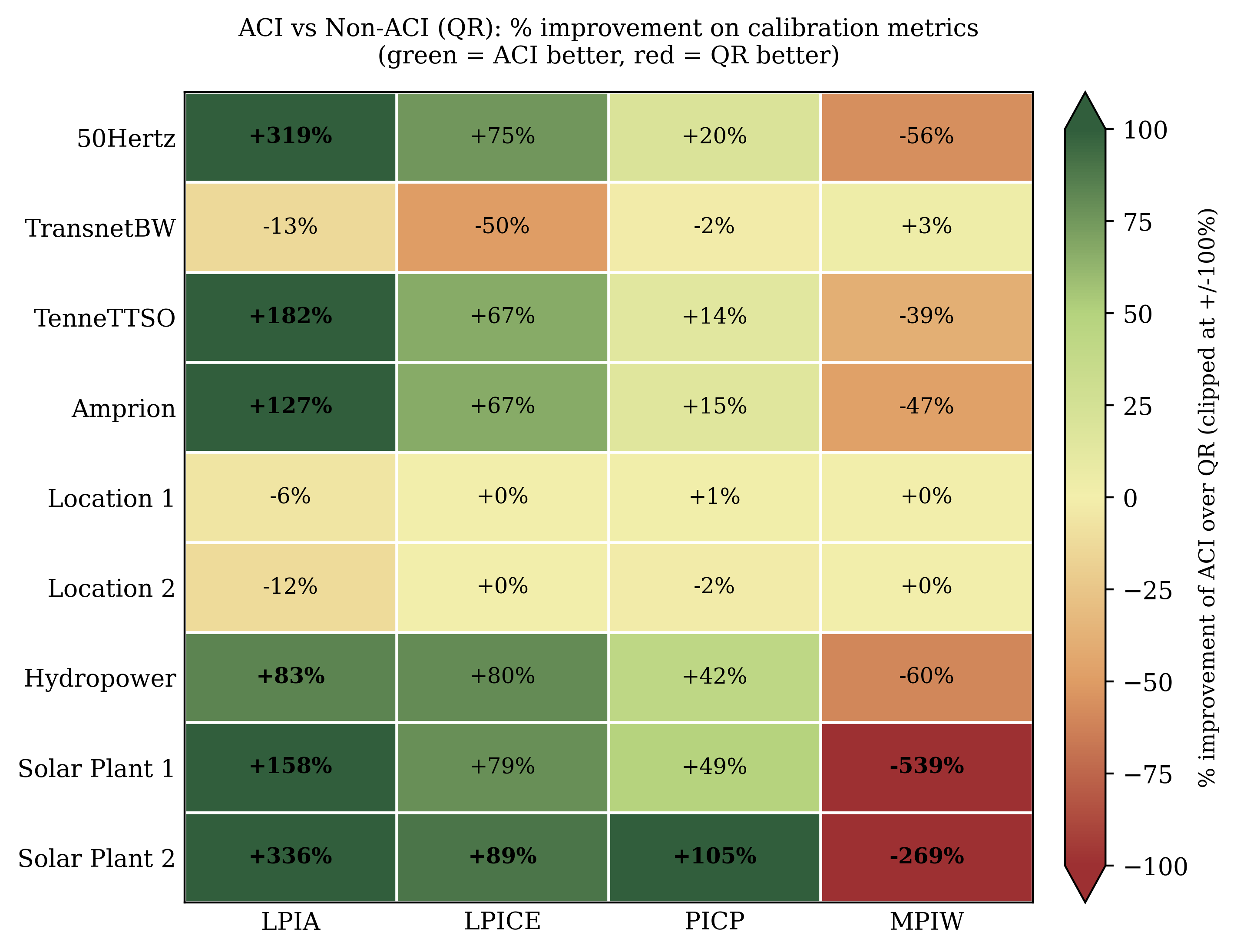}
        \caption{\% improvement}
        \label{fig:heatmap}
    \end{subfigure}
    \vspace{-2mm}
    \caption{ ACI performance improvement over the QR approach in terms of local calibration metrics defined in Section \ref{def_local_met}. (a) Distribution of Non-ACI QR (red) vs.\ ACI (blue) approach across datasets, showing ACI shifts LPICE lower, with tighter spread. (b) Heatmap of \% improvement of ACI over Non-ACI per dataset and metric, with ACI obtaining significantly positive improvement across most datasets and local metrics.}
    \label{fig:aci_vs_qr_all}
\end{figure}

Following the experimental design detailed in Section 6.1, we obtain the probabilistic forecasts using the LSTM model with the QR and ACI approaches across different energy datasets, as reported in Table \ref{tab:qr_aci_local_global}. We note that ACI achieves better local calibration than the QR approach across a substantial majority of the datasets in terms of the defined local metrics. Specifically, ACI achieves \textbf{higher PICP in 7 out of 9 datasets} and \textbf{higher LPIA in 6 out of 9 datasets}. For LPICE, where lower values indicate better performance, ACI outperforms QR in \textbf{6 out of 9 datasets}, while both approaches achieve identical values in the remaining two datasets. In Figure~\ref{fig:aci_vs_qr_all}(a), we compare the QR and ACI approaches in terms of LPICE and show that the ACI approach generally achieves lower LPICE than the QR approach. Further, the improvement in local calibration achieved by the ACI approach over the quantile regression approach is quantified in percentage across several local metrics defined in Section \ref{def_local_met} and plotted for each dataset in Figure~\ref{fig:aci_vs_qr_all}(b). Overall, these results empirically demonstrate that ACI improves local calibration compared with the QR approach across the evaluated energy datasets, making it a promising approach for risk-aware decision support in energy applications.

\subsection{Empirical Validation of the Assumption at Coverage Failures}
\label{subsec:summary_gain}

    To assess whether condition~(\ref{ass:severity-dominance}) holds in practice,
we apply ACI to the considered energy datasets and examine its local coverage
trajectory, in particular the frequency of abrupt increases in the local
coverage error, i.e.\ $\mathrm{LE}_s^{(L)} > \overline{\mathrm{LE}}_s^{(L)}$.
We follow the experimental setting described in
Section~\ref{subsec:multidomain_setup}: the first 5000 samples are reserved
for initial training, and ACI is then run online on top of an LSTM-based
probabilistic forecasting model, with $\gamma = 0.001$ and target coverage
$1-\alpha = 0.90$. The LSTM architecture and the tuned hyperparameters are
given in Table~\ref{tab:hyperparams} of Appendix~C. Our observations are
reported in Table~\ref{tab:laci_assumption_aci}. On an average around $62\%$ of observations suffers from the abrupt changes in local coverage errors. Also, across all datasets, the
local coverage error trajectory obeys
condition~(\ref{ass:severity-dominance}) with $p < 2$.

\begin{table}[H]
\centering
\resizebox{\textwidth}{!}{%
\begin{tabular}{lcccccc}
\toprule
\textbf{Dataset} &
\textbf{$\#(err_i=1)$} &
$\#(LE_{\hat{s}}^{(L)}>\overline{LE}_{\hat{s}} \mid err_i=1)$ &
\textbf{(\%)} &
\textbf{Mean $LE_{\hat{s}}^{(L)}$} &
\textbf{Mean $\overline{LE}_{\hat{s}}$} &
$\boldsymbol{\Delta}$ \\
\midrule

50Hertz
& 388 & 233 & 60.05 & 0.09 & 0.05 & 0.04 \\

Amprion
& 366 & 218 & 59.56 & 0.11 & 0.07 & 0.04 \\

Location1
& 482 & 285 & 59.13 & 0.08 & 0.05 & 0.04 \\

Location2
& 433 & 239 & 55.20 & 0.08 & 0.05 & 0.04 \\

Plant 1 Generation
& 231 & 139 & 60.17 & 0.22 & 0.19 & 0.02 \\

Plant 2 Generation
& 416 & 324 & 77.88 & 0.14 & 0.10 & 0.03 \\

TenneTTSO
& 401 & 238 & 59.35 & 0.10 & 0.06 & 0.04 \\

TransnetBW
& 420 & 241 & 57.38 & 0.06 & 0.03 & 0.03 \\

Hydropower
& 161 & 109 & 67.70 & 0.18 & 0.10 & 0.09 \\

\midrule
\textbf{Total}
& \textbf{3298}
& \textbf{2026}
& \textbf{61.43}
& \textbf{0.11}
& \textbf{0.07}
& \textbf{0.04} \\

\bottomrule
\end{tabular}%
}
\caption{Local coverage errors of ACI on the energy datasets. A large fraction
of the local coverage errors are abrupt, i.e.\
$\mathrm{LE}_s^{(L)} > \overline{\mathrm{LE}}_s^{(L)}$. On all datasets, the
local coverage error trajectory obeys
condition~(\ref{ass:severity-dominance}) with $p < 2$.}
\label{tab:laci_assumption_aci}

\end{table}

We further show that this behaviour of ACI does not depend on the choice of
$\gamma$. We repeat the experiment with different values of $\gamma$ and
report the results in
Tables~\ref{tab:local_error_gamma_004}--\ref{tab:local_error_gamma_016} of
Appendix~\ref{app:local_error_assumption_different_ACI}.
Condition~(\ref{ass:severity-dominance})  holds with $p<2$ across all values of $\gamma$
considered.




\subsection{ LACI leads to better local calibration in energy forecasting tasks}
\label{subsec:error_severity}

\begin{table*}[t]
\centering
\scriptsize
\setlength{\tabcolsep}{4.5pt}
\renewcommand{\arraystretch}{1.05}

\begin{tabular}{ll|ccc|cc}
\toprule
\textbf{Dataset} & \textbf{Method} &
\multicolumn{3}{c|}{\textbf{Local Metrics}} &
\multicolumn{2}{c}{\textbf{Global Metrics}} \\
\cmidrule(lr){3-5}
\cmidrule(lr){6-7}
& &
\textbf{LPICE} &
\textbf{LPIA} &
\textbf{LMPIW} &
\textbf{PICP} &
\textbf{MPIW} \\
\midrule

\multirow{5}{*}{50Hertz}
& ACI          & 0.04 & 0.67 & 49.10 & 0.90 & 49.12 \\
& Weighted ACI & 0.04 & 0.63 & 48.87 & 0.90 & 48.88 \\
& LACI         & \textbf{0.03} & \textbf{0.73} & 55.09 & \textbf{0.93} & 55.09 \\
& DtACI        & 0.04 & 0.63 & \textbf{48.09} & 0.90 & \textbf{48.09} \\
& AgACI        & 0.04 & 0.63 & 49.63 & 0.90 & 49.64 \\
\midrule

\multirow{5}{*}{Amprion}
& ACI          & 0.04 & 0.68 & 32.03 & 0.91 & 32.05 \\
& Weighted ACI & 0.04 & 0.66 & \textbf{31.83} & 0.91 & 31.85 \\
& LACI         & \textbf{0.03} & \textbf{0.75} & 36.33 & 0.93 & 36.33 \\
& DtACI        & 0.04 & 0.66 & 30.88 & 0.90 & 30.89 \\
& AgACI        & 0.04 & 0.67 & 34.16 & 0.91 & 34.15 \\
\midrule

\multirow{5}{*}{TenneTTSO}
& ACI          & 0.04 & 0.62 & \textbf{39.99} & 0.90 & 39.99 \\
& Weighted ACI & 0.04 & 0.62 & 42.06 & 0.90 & 42.04 \\
& LACI         & \textbf{0.03} & \textbf{0.71} & 44.44 & 0.92 & 44.43 \\
& DtACI        & 0.04 & 0.59 & 39.60 & 0.90 & 39.60 \\
& AgACI        & 0.04 & 0.60 & 43.06 & 0.90 & 43.04 \\
\midrule

\multirow{5}{*}{TransnetBW}
& ACI          & 0.03 & 0.62 & 11.82 & 0.90 & 11.81 \\
& Weighted ACI & 0.03 & 0.62 & 11.92 & 0.90 & 11.91 \\
& LACI         & \textbf{0.01} & \textbf{0.80} & 13.28 & 0.94 & 13.26 \\
& DtACI        & 0.03 & 0.62 & \textbf{11.79} & 0.90 & 11.78 \\
& AgACI        & 0.03 & 0.61 & 12.79 & 0.90 & 12.77 \\
\midrule

\multirow{5}{*}{Hydropower}
& ACI          & 0.06 & 0.54 & 202.22 & 0.88 & 202.91 \\
& Weighted ACI & 0.07 & 0.58 & \textbf{147.09} & 0.87 & 147.11 \\
& LACI         & 0.05 & 0.59 & 211.04 & 0.90 & 211.60 \\
& DtACI        & \textbf{0.04} & 0.54 & 192.49 & 0.90 & 193.30 \\
& AgACI        & 0.05 & \textbf{0.60} & 204.11 & 0.90 & 204.76 \\
\midrule

\multirow{5}{*}{Location 1}
& ACI          & 0.03 & 0.58 & 0.28 & 0.90 & 0.28 \\
& Weighted ACI & 0.04 & 0.62 & \textbf{0.03} & 0.90 & 0.03 \\
& LACI         & \textbf{0.02} & \textbf{0.74} & 0.33 & 0.93 & 0.33 \\
& DtACI        & 0.03 & 0.55 & 0.28 & 0.90 & 0.28 \\
& AgACI        & 0.03 & 0.57 & 0.28 & 0.90 & 0.28 \\
\midrule

\multirow{5}{*}{Location 2}
& ACI          & 0.04 & 0.55 & 0.20 & 0.89 & 0.20 \\
& Weighted ACI & 0.04 & 0.61 & \textbf{0.03} & 0.90 & 0.03 \\
& LACI         & \textbf{0.01} & \textbf{0.76} & 0.25 & 0.93 & 0.25 \\
& DtACI        & 0.03 & 0.56 & 0.20 & 0.90 & 0.20 \\
& AgACI        & 0.03 & 0.56 & 0.21 & 0.90 & 0.21 \\
\midrule

\multirow{5}{*}{Plant 1}
& ACI          & 0.05 & 0.66 & \textbf{749.16} & 0.90 & 751.75 \\
& Weighted ACI & 0.07 & 0.62 & 1435.14 & 0.88 & 1434.73 \\
& LACI         & 0.04 & \textbf{0.73} & 837.24 & 0.92 & 840.13 \\
& DtACI        & 0.04 & 0.60 & 687.07 & 0.90 & 688.99 \\
& AgACI        & 0.04 & 0.60 & 886.86 & 0.91 & 888.70 \\
\midrule

\multirow{5}{*}{Plant 2}
& ACI          & 0.05 & 0.60 & 4752.39 & 0.89 & 4755.29 \\
& Weighted ACI & 0.05 & 0.41 & 5630.51 & 0.88 & 5637.40 \\
& LACI         & 0.03 & \textbf{0.75} & 7462.30 & 0.93 & 7456.76 \\
& DtACI        & 0.03 & 0.56 & \textbf{4534.31} & 0.90 & 4541.26 \\
& AgACI        & 0.03 & 0.61 & 5047.83 & 0.90 & 5051.95 \\

\bottomrule
\end{tabular}

\caption{Comparison of ACI, Weighted ACI, LACI, DtACI,
and AgACI across nine energy forecasting datasets at
nominal coverage $1-\alpha=0.90$. LACI achieves the
highest LPIA in 8 out of 9 datasets and the lowest LPICE
in 8 out of 9 datasets, demonstrating improved local
calibration performance. Best results are highlighted
in bold.}
\label{tab:all_methods_comparison_LSTM}
\end{table*}

\begin{table*}[t]
\centering

\scriptsize
\setlength{\tabcolsep}{5pt}
\renewcommand{\arraystretch}{1.05}
\begin{tabular}{ll|ccc|cc}
\toprule
\textbf{Dataset} & \textbf{Method} &
\multicolumn{3}{c|}{\textbf{Local Metrics}} &
\multicolumn{2}{c}{\textbf{Global Metrics}} \\
\cmidrule(lr){3-5} \cmidrule(lr){6-7}
& &
\textbf{LPIA} & \textbf{LPICE} & \textbf{LMPIW} &
\textbf{PICP} & \textbf{MPIW} \\
\midrule

\multirow{4}{*}{50Hertz}
& ACI   & 0.62 & 0.04 & 49.55 & 0.90 & 49.57 \\
& LACI  & \textbf{0.74} & \textbf{0.02} & 57.05 & \textbf{0.93} & 57.05 \\
& DtACI & 0.58 & 0.04 & \textbf{48.91} & 0.90 & \textbf{48.91} \\
& AgACI & 0.57 & 0.04 & 49.43 & 0.90 & 49.44 \\
\midrule

\multirow{4}{*}{Amprion}
& ACI   & 0.66 & 0.04 & 32.47 & 0.91 & 32.49 \\
& LACI  & \textbf{0.73} & \textbf{0.03} & 36.40 & \textbf{0.93} & 36.42 \\
& DtACI & 0.65 & 0.04 & \textbf{30.56} & 0.90 & \textbf{30.59} \\
& AgACI & 0.62 & 0.04 & 32.97 & 0.90 & 32.98 \\
\midrule

\multirow{4}{*}{TransnetBW}
& ACI   & 0.66 & 0.03 & 12.36 & 0.90 & 12.34 \\
& LACI  & \textbf{0.78} & \textbf{0.02} & 13.47 & \textbf{0.93} & 13.45 \\
& DtACI & 0.62 & 0.03 & \textbf{12.20} & \textbf{0.90} & \textbf{12.18} \\
& AgACI & 0.62 & 0.03 & 12.83 & \textbf{0.90} & 12.81 \\
\midrule

\multirow{4}{*}{TenneTTSO}
& ACI   & 0.62 & 0.04 & 43.58 & 0.90 & 43.55 \\
& LACI  & \textbf{0.74} & \textbf{0.02} & 50.34 & \textbf{0.94} & 50.28 \\
& DtACI & 0.59 & 0.04 & \textbf{42.27} & 0.90 & \textbf{42.24} \\
& AgACI & 0.58 & 0.04 & 42.45 & 0.90 & 42.42 \\
\midrule

\multirow{4}{*}{Location 1}
& ACI   & 0.56 & 0.03 & 0.26 & 0.90 & 0.26 \\
& LACI  & \textbf{0.79} & \textbf{0.01} & 0.33 & \textbf{0.93} & 0.33 \\
& DtACI & 0.54 & 0.03 & \textbf{0.26} & 0.90 & \textbf{0.26} \\
& AgACI & 0.57 & 0.03 & 0.26 & 0.90 & 0.26 \\
\midrule

\multirow{4}{*}{Location 2}
& ACI   & 0.55 & 0.04 & \textbf{0.19} & 0.89 & \textbf{0.19} \\
& LACI  & \textbf{0.74} & \textbf{0.02} & 0.24 & \textbf{0.93} & 0.24 \\
& DtACI & 0.53 & 0.03 & 0.19 & 0.90 & 0.19 \\
& AgACI & 0.56 & 0.03 & 0.20 & 0.90 & 0.20 \\
\midrule

\multirow{4}{*}{Solar Plant 1}
& ACI   & 0.66 & 0.06 & \textbf{628.20} & 0.90 & \textbf{627.59} \\
& LACI  & \textbf{0.72} & 0.04 & 910.52 & \textbf{0.92} & 909.01 \\
& DtACI & 0.58 & 0.04 & 634.90 & 0.90 & 633.71 \\
& AgACI & 0.63 & \textbf{0.04} & 795.57 & 0.91 & 793.95 \\
\midrule

\multirow{4}{*}{Solar Plant 2}
& ACI   & 0.56 & 0.05 & 5226.78 & 0.89 & 5230.68 \\
& LACI  & \textbf{0.71} & \textbf{0.02} & 6783.43 & \textbf{0.93} & 6785.30 \\
& DtACI & 0.58 & 0.03 & \textbf{4895.05} & 0.90 & \textbf{4905.16} \\
& AgACI & 0.56 & 0.04 & 5370.65 & 0.90 & 5379.70 \\
\midrule

\multirow{4}{*}{Hydropower}
& ACI   & 0.66 & 0.07 & 150.97 & 0.87 & 151.49 \\
& LACI  & \textbf{0.70} & 0.06 & 157.72 & 0.89 & 158.15 \\
& DtACI & 0.60 & 0.04 & \textbf{141.34} & 0.90 & \textbf{142.04} \\
& AgACI & 0.65 & \textbf{0.04} & 172.13 & \textbf{0.91} & 172.31 \\

\bottomrule
\end{tabular}

\caption{Comparison of ACI, LACI, DtACI, and AgACI across nine energy forecasting datasets using a GRU architecture at a nominal coverage level of $1-\alpha=0.90$. LACI achieves the highest LPIA across all nine datasets and the lowest LPICE in 7 out 9 datasets, demonstrating consistently improved local calibration performance.  The best results for each metric are highlighted in bold.}

\label{tab:aci_laci_dtaci_agaci_all_GRU}
\end{table*}

\begin{table*}[t]
\centering

\scriptsize
\setlength{\tabcolsep}{5pt}
\renewcommand{\arraystretch}{1.05}
\begin{tabular}{ll|ccc|cc}
\toprule
\textbf{Dataset} & \textbf{Method} &
\multicolumn{3}{c|}{\textbf{Local Metrics}} &
\multicolumn{2}{c}{\textbf{Global Metrics}} \\
\cmidrule(lr){3-5} \cmidrule(lr){6-7}
& &
\textbf{LPIA} & \textbf{LPICE} & \textbf{LMPIW} &
\textbf{PICP} & \textbf{MPIW} \\
\midrule

\multirow{4}{*}{50Hertz}
& ACI   & 0.55 & 0.03 & 43.07 & 0.90 & 43.05 \\
& LACI  & \textbf{0.76} & \textbf{0.01} & 49.91 & 0.93 & 49.91 \\
& DtACI & 0.55 & 0.03 & \textbf{42.99} & \textbf{0.90} & \textbf{42.97} \\
& AgACI & 0.56 & 0.03 & 43.15 & 0.90 & 43.14 \\

\midrule

\multirow{4}{*}{Amprion}
& ACI   & 0.59 & 0.03 & \textbf{25.85} & 0.90 & \textbf{25.85} \\
& LACI  & \textbf{0.75} & \textbf{0.02} & 29.56 & 0.93 & 29.55 \\
& DtACI & 0.58 & 0.03 & 26.08 & \textbf{0.90} & 26.09 \\
& AgACI & 0.59 & 0.03 & 26.16 & 0.90 & 26.16 \\

\midrule

\multirow{4}{*}{TenneTTSO}
& ACI   & 0.59 & 0.03 & 36.54 & \textbf{0.90} & 36.50 \\
& LACI  & \textbf{0.77} & \textbf{0.01} & 41.43 & 0.94 & 41.38 \\
& DtACI & 0.59 & 0.03 & 36.05 & 0.90 & 36.01 \\
& AgACI & 0.58 & 0.03 & \textbf{36.05} & 0.90 & \textbf{36.01} \\

\midrule

\multirow{4}{*}{TransnetBW}
& ACI   & 0.61 & 0.03 & \textbf{11.24} & 0.90 & \textbf{11.23} \\
& LACI  & \textbf{0.78} & \textbf{0.01} & 12.22 & 0.93 & 12.21 \\
& DtACI & 0.57 & 0.03 & 11.28 & \textbf{0.90} & 11.26 \\
& AgACI & 0.58 & 0.03 & 11.30 & 0.90 & 11.29 \\

\midrule

\multirow{4}{*}{Location 1}
& ACI   & 0.57 & 0.03 & \textbf{0.26} & 0.90 & \textbf{0.26} \\
& LACI  & \textbf{0.81} & \textbf{0.01} & 0.32 & 0.94 & 0.32 \\
& DtACI & 0.54 & 0.02 & 0.26 & \textbf{0.90} & 0.26 \\
& AgACI & 0.56 & 0.03 & 0.26 & 0.90 & 0.26 \\

\midrule

\multirow{4}{*}{Location 2}
& ACI   & 0.54 & 0.03 & \textbf{0.18} & 0.90 & \textbf{0.18} \\
& LACI  & \textbf{0.74} & \textbf{0.01} & 0.21 & 0.93 & 0.21 \\
& DtACI & 0.54 & 0.02 & 0.18 & \textbf{0.90} & 0.18 \\
& AgACI & 0.55 & 0.03 & 0.18 & 0.90 & 0.18 \\

\midrule

\multirow{4}{*}{Solar Plant 1}
& ACI   & 0.60 & 0.04 & \textbf{564.00} & 0.90 & \textbf{562.88} \\
& LACI  & \textbf{0.78} & \textbf{0.02} & 678.77 & 0.94 & 677.37 \\
& DtACI & 0.58 & 0.04 & 586.90 & \textbf{0.90} & 585.86 \\
& AgACI & 0.63 & 0.04 & 604.85 & 0.91 & 603.58 \\

\midrule

\multirow{4}{*}{Solar Plant 2}
& ACI   & 0.59 & 0.06 & 4789.80 & 0.89 & 4799.47 \\
& LACI  & \textbf{0.76} & \textbf{0.02} & 7224.34 & 0.94 & 7227.63 \\
& DtACI & 0.58 & 0.03 & \textbf{4593.01} & \textbf{0.90} & \textbf{4601.81} \\
& AgACI & 0.58 & 0.03 & 5199.46 & 0.91 & 5208.07 \\

\midrule

\multirow{4}{*}{Hydropower}
& ACI   & 0.57 & 0.05 & 134.14 & 0.88 & 135.58 \\
& LACI  & \textbf{0.62} & \textbf{0.04} & 146.56 & 0.91 & 148.10 \\
& DtACI & 0.53 & 0.04 & \textbf{130.12} & 0.89 & \textbf{131.60} \\
& AgACI & 0.59 & 0.04 & 137.62 & \textbf{0.90} & 139.23 \\

\bottomrule
\end{tabular}

\caption{Comparison of ACI, LACI, DtACI, and AgACI across nine energy forecasting datasets using  TCN Architecture  at nominal coverage ($1-\alpha=0.90$). LACI consistently achieves the highest LPIA and lowest LPICE across all nine datasets, demonstrating superior local calibration performance.  Best results are highlighted in bold.}

\label{tab:aci_laci_dtaci_agaci_all_TCN}
\end{table*}

We now present the numerical experiments to empirically verify our claim that LACI can provide improved local calibration for energy forecasting tasks.

In selecting the baselines, we note that LACI is a modification of the ACI
recursion. We therefore compare it first with the standard ACI and its
recency-weighted variant, which differ from LACI only in how past errors are
weighted. We also include two recent ACI variants, DtACI
\citep{gibbs2024conformal} and AgACI \citep{pmlr-v162-zaffran22a}, which
retain the ACI recursion but aggregate multiple ACI instances with different
step sizes through online expert aggregation. Although LACI uses a single
recursion and no expert aggregation, we include these more sophisticated
variants to assess local calibration capability against the strongest methods of the same type. 

We estrict our comparison to methods that adapt the miscoverage level
$\alpha_i$ through the ACI recursion and exclude online conformal methods that track the interval threshold
directly in the space of nonconformity scores, such as Conformal SAOCP \citep{bhatnagar2023improved},
instead of using the ACI recursion. Since LACI is a simple modification of the
standard ACI update, restricting the comparison to this class isolates the
effect of its local-error weighting.

Table~\ref{tab:all_methods_comparison_LSTM} compares ACI, LACI, DtACI, and AgACI, all trained with the base LSTM architecture with the pinball loss. ACI and Weighted ACI recrusion obtains the simmilar perfromance across aall  While the existing ACI variants perform reasonably well on global metrics such as PICP and MPIW, they tend to perform poorly at the local level. The average {LPIA} obtained by ACI, DtACI, and AgACI is $0.61$, $0.59$, and $0.60$, respectively, meaning that these methods achieve the target coverage across local windows only $61\%$, $59\%$, and $60\%$ of the time.  In contrast, the LACI variant raises this figure to $\textbf{72\%}$ along with the best \textbf{LPIA} in $\mathbf{8}$ \textbf{out of} $\mathbf{9}$ cases, at a modest cost in interval width as reflected by the LMPIW values, thereby improving the overall local calibration quality of the model. Additionally, LACI achieves the lowest {LPICE} in 8 of 9 datasets. The LPICE metric characterizes the average magnitude of local calibration error.

\begin{figure}[H]
\centering
\begin{subfigure}[b]{0.48\columnwidth}
    \centering
    \includegraphics[width=\textwidth]{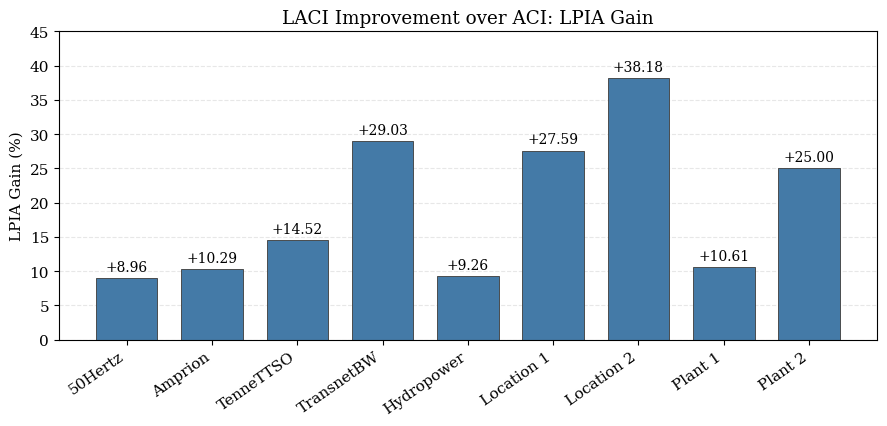}
    \caption{LPIA Gain (\%)}
    \label{fig:lpia_gain}
\end{subfigure}
\hfill
\begin{subfigure}[b]{0.48\columnwidth}
    \centering
    \includegraphics[width=\textwidth]{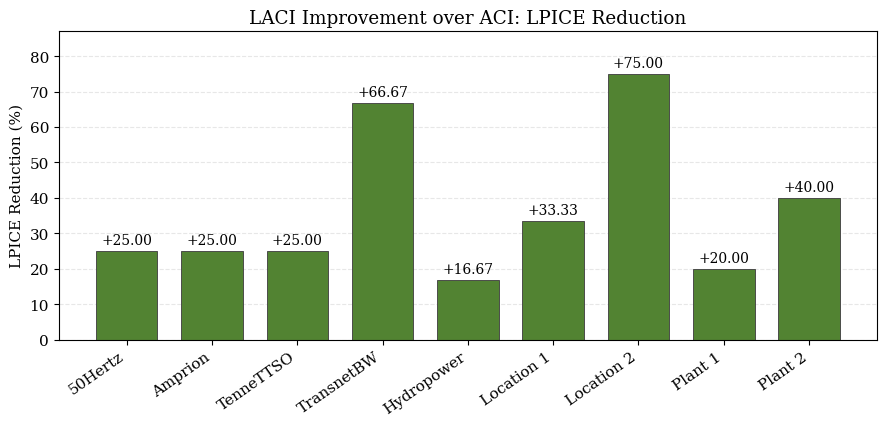}
    \caption{LPICE Reduction (\%)}
    \label{fig:lpice_reduction}
\end{subfigure}
\caption{Percentage improvement of LACI over ACI in local calibration metrics. (a) LACI yields consistent gains in LPIA and  (b) reductions in LPICE errors across most datasets.}
\label{fig:lpia_lpice_gain}
\end{figure}

Further, we quantify the improvement achieved by LACI over ACI in terms of LPIA and LPICE. Figure~\ref{fig:lpia_gain} visualizes the percentage improvement in LPIA obtained by LACI over ACI, while Figure~\ref{fig:lpice_reduction} visualizes the percentage reduction in LPICE obtained by LACI over ACI, for each dataset. For TransnetBW, Location~1, and Location~2, LACI achieves a particularly significant improvement in local calibration ability over ACI.

The advantages of LACI over ACI and its existing variants are independent of the neural architecture used to obtain the probabilistic forecasts. To verify this, we repeat  above experiments with different neural architectures, namely GRU and TCN. Tables~\ref{tab:aci_laci_dtaci_agaci_all_GRU} and~\ref{tab:aci_laci_dtaci_agaci_all_TCN}  compare the quality of the probabilistic forecasts at the local and global level obtained by LACI against ACI and its existing variants, trained using the GRU and TCN architectures, respectively. Across both GRU and TCN architectures, LACI consistently outperforms ACI and its variants in terms of local calibration. Specifically, LACI achieves the highest \textbf{LPIA in all 9 out of 9 cases} for both architectures, demonstrating its effectiveness in capturing local variations in forecast uncertainty. It also generally achieves the lowest \textbf{LPICE}, indicating improved local calibration accuracy. These results highlight the robustness and effectiveness of LACI across different forecasting architectures. The neural architectures and tuned hyperparameters used in these experiments
are listed in Table~\ref{tab:hyperparams} of Appendix C.


\section{Case study:  Economic implications of  improvement of Local Calibration}
\label{sec:case_study}

In electricity markets, power generation companies are required to schedule
their expected electricity generation in advance. When the actual generation
deviates from the scheduled generation, the resulting imbalance is settled
through an imbalance pricing mechanism, which can lead to financial penalties
or additional settlement costs for the generator. In India, such deviations
are governed by the Deviation Settlement Mechanism (DSM) regulations issued by
the Central Electricity Regulatory Commission (CERC), while Germany uses the
uniform imbalance price mechanism (reBAP) for balancing-group settlement.
These mechanisms help market participants plan their energy portfolios in
advance, encourage accurate generation scheduling, and help maintain the
stability of the power grid.

We consider a simple hourly revenue model for the power generation company,
$R_t = p_t^{(g)} E_t^{(g)} - C_t^{\mathrm{imb}}$,
where $R_t$ denotes the net revenue at hour $t$, $p_t^{(g)}$ is the selling
price per unit of electricity generated at hour $t$, $E_t^{(g)}$ represents
the total units of electricity generated at hour $t$, and
$C_t^{\mathrm{imb}}$ is the imbalance penalty cost at hour $t$, which is
charged when the power company fails to meet its commitment.

High-quality PI estimation can help the power plant manager project future
generation outcomes with high reliability, unlike a single point estimate,
which provides no quantification of the prediction error and may lead to
heavy penalty costs. To estimate such intervals, an online conformal model is
deployed, which generates hourly rolling forecasts of the power interval with
target coverage $1-\alpha$. From the client's perspective, it is important
that the power company meets its supply commitment over each 24-hour
operational window, rather than only over a long time horizon. This requires
the probabilistic forecasting model to yield power intervals that are locally
well calibrated over the 24-hour operational window. In the simplest setting,
we consider the following imbalance penalty cost:
\begin{equation}
  C_i^{\mathrm{imb}} = P_{\mathrm{mis}}\,\mathrm{LE}_i + \lambda\,\mathrm{LMPIW}_i ,
  \label{func_cost_im}
\end{equation}
where $P_{\mathrm{mis}}$ represents the penalty coefficient and
$\mathrm{LE}_i$ is the local coverage error at the $i$-th hour, which
quantifies the violation of the power company's energy commitment within the
24-hour operational window. In addition, the width of the local energy
interval is penalised at a lower rate through the coefficient $\lambda$,
which represents the cost of holding flexible reserve to cover the
uncertainty band. The average penalty over the evaluation period
$t+1, t+2, \ldots, t+k$ is given by
\begin{equation}
  \overline{C}
  = P_{\mathrm{mis}}\,\frac{1}{k}\sum_{i=t+1}^{t+k}\mathrm{LE}_i
  + \lambda\,\frac{1}{k}\sum_{i=t+1}^{t+k}\mathrm{LMPIW}_i
  = P_{\mathrm{mis}}\,\mathrm{LPICE} + \lambda\,\mathrm{LMPIW}.
  \label{eq:case_mean_cost}
\end{equation}

The validity of our cost model~(\ref{func_cost_im}) is supported by the
equivalent-cost approach to reserve setting under probabilistic wind power
forecasts \citep{matos2011setting}, which likewise trades the cost of
flexibility against a reliability risk measure through a constant trade-off
coefficient, examined over a range of values. Our model adopts this
structure, with the interval width representing the cost of flexibility and
the local coverage error over operating windows serving as the risk measure.

For the economic evaluation, we conduct rolling forecasting experiments using the considered conformal prediction methods, namely ACI, DtACI, AgACI, and the proposed LACI. During the forecasting process, the prediction intervals generated by each method are evaluated against the corresponding observed values. We then compute the local prediction interval coverage error (LPICE) and the local mean prediction interval width (LMPIW), aggregated over the evaluation period for each method and dataset. Finally, these aggregated metrics are substituted into the composite cost function defined in ~\ref{eq:case_mean_cost} to quantify the overall cost associated with each method.

In the aforementioned experimental setup, the parameter values are set to $P_{\mathrm{mis}} = \text{\texteuro}\,200$ per unit of local coverage error and $\lambda =\text{\texteuro}$ 0.1\, per unit of prediction interval width. We evaluate all the considered ACI variants,
including our LACI model, and compute the overall penalty cost incurred by
their deployment. The results, reported in
Table~\ref{tab:local_calibration_cost}, show that LACI yields the lowest
imbalance cost in $5$ of the $8$ evaluated cases.

\begin{table*}[t]
\centering
\scriptsize
\setlength{\tabcolsep}{4.5pt}
\renewcommand{\arraystretch}{1.05}

\begin{tabular}{llcccc}
\toprule
\textbf{Dataset} & \textbf{Method} &
\textbf{LPICE} $\downarrow$ &
\textbf{LPIA} $\uparrow$ &
\textbf{LMPIW} $\downarrow$ &
\textbf{Total Composite Cost(in $\text{\texteuro}$)} $\downarrow$ \\
\midrule

\multirow{4}{*}{50Hertz}
& ACI   & 0.04 & 0.67 & 49.14 & 57354.14 \\
& DtACI & 0.04 & 0.63 & \textbf{48.10} & 56087.83 \\
& AgACI & 0.04 & 0.63 & 49.64 & 54639.35 \\
& LACI  & \textbf{0.03} & \textbf{0.73} & 55.12 & \textbf{48062.00} \\
\midrule

\multirow{4}{*}{Amprion}
& ACI   & 0.04 & 0.68 & 32.07 & 51740.82 \\
& DtACI & 0.04 & 0.66 & \textbf{30.89} & 48858.61 \\
& AgACI & 0.04 & 0.67 & 34.17 & 46768.42 \\
& LACI  & \textbf{0.03} & \textbf{0.75} & 36.35 & \textbf{41998.42} \\
\midrule

\multirow{4}{*}{Location 1}
& ACI   & 0.03 & 0.58 & 0.28 & 37342.30 \\
& DtACI & 0.03 & 0.55 & \textbf{0.28} & 33110.13 \\
& AgACI & 0.03 & 0.57 & 0.28 & 33317.04 \\
& LACI  & \textbf{0.02} & \textbf{0.74} & 0.33 & \textbf{18381.23} \\
\midrule

\multirow{4}{*}{Location 2}
& ACI   & 0.04 & 0.55 & 0.20 & 41020.21 \\
& DtACI & 0.03 & 0.56 & \textbf{0.20} & 32614.99 \\
& AgACI & 0.03 & 0.56 & 0.21 & 34885.78 \\
& LACI  & \textbf{0.01} & \textbf{0.76} & 0.25 & \textbf{17570.54} \\
\midrule

\multirow{4}{*}{Plant 1}
& ACI   & 0.05 & 0.66 & 749.16 & 503939.92 \\
& DtACI & 0.04 & 0.60 & \textbf{687.07} & \textbf{451546.70} \\
& AgACI & \textbf{0.04} & 0.60 & 886.86 & 566952.84 \\
& LACI  & 0.04 & \textbf{0.73} & 837.24 & 539169.39 \\
\midrule

\multirow{4}{*}{TenneTTSO}
& ACI   & 0.04 & 0.62 & 40.00 & 53651.14 \\
& DtACI & 0.04 & 0.59 & \textbf{39.61} & 52912.20 \\
& AgACI & 0.04 & 0.60 & 43.06 & 51058.34 \\
& LACI  & \textbf{0.03} & \textbf{0.71} & 44.45 & \textbf{45777.83} \\
\midrule

\multirow{4}{*}{TransnetBW}
& ACI   & 0.03 & 0.62 & 11.82 & 31541.34 \\
& DtACI & 0.03 & 0.62 & \textbf{11.79} & 29818.32 \\
& AgACI & 0.03 & 0.61 & 12.79 & 29314.38 \\
& LACI  & \textbf{0.01} & \textbf{0.80} & 13.28 & \textbf{17734.68} \\
\midrule

\multirow{4}{*}{Hydropower}
& ACI   & 0.06 & 0.54 & 202.22 & 47304.91 \\
& DtACI & \textbf{0.04} & 0.54 & \textbf{192.49} & \textbf{38839.77} \\
& AgACI & 0.05 & \textbf{0.60} & 204.11 & 42818.17 \\
& LACI  & 0.05 & 0.59 & 211.04 & 44273.67 \\

\bottomrule
\end{tabular}
\caption{Total Composite Cost(in $\text{\texteuro}$) , computed using (~\ref{eq:case_mean_cost}) for ACI and its variants, including LACI, with
$P_{\mathrm{mis}} = \text{\texteuro}\,200$ per unit of local coverage error and $\lambda =\text{\texteuro}$ 0.1\, per unit of prediction interval width. LACI achieves better local
calibration and thus a lower imbalance penalty cost in $5$ of the $8$
evaluated cases.} 
\label{tab:local_calibration_cost}
\end{table*}

\section{Limitation and Future Works}
\label{sec:conclusion}

The findings of this study highlight the need to move beyond global
calibration towards locally reliable  intervals, particularly in
operational settings where decisions are made over short time windows.  Achieving the desired coverage over a long
evaluation period does not necessarily ensure reliable predictions during
individual operating periods. This motivates the development of online
conformal methods that explicitly account for local calibration requirements
while maintaining narrow intervals.

The local calibration requirement plays a crucial role in decision support
across applications. However, how a locally calibrated probabilistic forecast
should be translated into risk-aware decisions is inherently application
dependent: the relevant operating window, the relative cost of under- and
over-coverage, and the cost of wider intervals differ across domains. A natural next step is to develop application-specific local
evaluation metrics derived from the actual decision process, and online
probabilistic forecasting methods that directly minimize them, rather than a
generic coverage target.

LACI improves the local calibration of standard ACI and its variants, but it
is deliberately simple.  An important direction is to
combine LACI's local-error weighting with recent online conformal methods that
aggregate multiple experts or online learners, such as DtACI
\citep{gibbs2024conformal} and AgACI \citep{pmlr-v162-zaffran22a}. These methods
adapt to changing distributions by combining experts with different adaptation
rates. Incorporating local coverage errors into such frameworks may allow the
intervals to respond more effectively to abrupt shifts while balancing local
coverage and interval width.

We showed that ACI and weighted ACI variant attain the GCR, and that LACI empirically improves local calibration of weighted ACI. However, we have not established explicit global and local guarantees for LACI
itself. Future work should derive theoretical results in this direction.

\section*{Funding}
This work was funded by the Smart Energy Learning Center at Dhirubhai Ambani University (Formerly
DA-IICT), Gandhinagar through grant no.- CSR-25/BSES/A6-PA/SELC.

 \bibliography{cite} 
\bibliographystyle{tmlr}

\appendix

\section{Proofs of Theoretical Results}
In this appendix, we provide the proofs of all lemmas and propositions stated
in the main body of the paper. In addition, we analyse the weighted ACI
process and prove that it attains the GCR.

\subsection{Proof of Proposition ~\ref{prop:lpicp_identity}}
\label{appendix:proof_lpicp}

\begin{proof}
Recall the ACI update rule:
\begin{equation}
    \alpha_{{\hat{i}}+1} = \alpha_{\hat{i}} + \gamma(\alpha - \mathrm{err}_{\hat{i}})
    \label{eq:aci_update}
\end{equation}
where $\mathrm{err}_{\hat{i}} = 1 - z(h_i)$ and $z(h_i) := \mathbb{I}\{x_{i+1} \in \hat{C}^{(\alpha_{\hat{i}})} (h_i)\}$ 
is the binary coverage indicator at $i^{th}$ time-series observation ($\hat{i}^{th}$ data stream after the base model is trained with $T_{int}$ samples).
Substituting $\mathrm{err}_{\hat{i}} = 1 - z(h_i)$ in ACI process (\ref{eq:aci_update}) gives 
\begin{equation*}
    \alpha_{{\hat{i}}+1} 
    = \alpha_{\hat{i}} + \gamma\bigl(\alpha - (1 - z(h_i))\bigr)
    = \alpha_{\hat{i}} - \gamma\bigl((1-\alpha) - z(h_i)\bigr).
    \label{eq:aci_expanded}
\end{equation*}
Rearranging the above yields an explicit expression for $z(h_i)$:
\begin{equation}
    z(h_i) = (1-\alpha) + \frac{\alpha_{{\hat{i}}+1} - \alpha_{\hat{i}}}{\gamma}.
    \label{eq:z_decomposition}
\end{equation}

Consider the choice of window $  \mathcal{N}_w^{L}(h_i) = \{ h_j \mid j = i-w+1, \ldots, i \}$  of size $w$, then summing (\ref{eq:z_decomposition}) over $\mathcal{N}_w^{L}(h_i)$ yields
\begin{equation}
    \sum_{ j \in \mathcal{N}_w^{L}(h_i) } z_j 
    = w(1-\alpha) 
    + \frac{1}{\gamma} \sum_{{j} \in \mathcal{N}_w^{L}(h_i) } (\alpha_{j+1} - \alpha_j)=  w(1-\alpha)   + (\alpha_{\hat{i}+1} - \alpha_{\hat{i}-w+1})
    \label{eq:sum_z}
\end{equation}

Noting that $z_i := \mathbb{I}\{x_{i+1} \in \hat{C}^{(\alpha_i)} (h_i)\}$ denotes the belonging of the $y_i$ in estimated conformal interval, (\ref{eq:sum_z}) reduces in view of  (\ref{lpicp}) as
\begin{equation}
    \mathrm{LPICP}^{(L)}_{\hat{i}} 
    = (1-\alpha) + \frac{\Delta\alpha_{\hat{i}}}{\gamma w},
    \label{eq:lpicp_derived}
\end{equation}
where $\Delta\alpha_{\hat{i}}$ denotes the overall evolution of $\alpha$ values across the window  $\mathcal{N}_w^{L}(h_i)$ that further implies
\begin{equation*}
    \mathrm{LE}^{(L)}_{\hat{i}}
    = \max\!\bigl((1-\alpha) - \mathrm{LPICP}^{(L)}_{\hat{i}},\; 0\bigr) 
    = \max\!\left(-\frac{\Delta\alpha_{\hat{i}}}{\gamma w},\; 0\right),
\end{equation*}
and completes the proof.
\end{proof}

\subsection{Proof of Lemma 1}

We note that the $\{\mathrm{err}_{\hat{s}}\}_{\hat{s}=1}^{\hat{i}}$, $\mathrm{err}_{\hat{s}} \in\{0,1\}$ is the  fixed binary coverage-error trajectory generated by ACI process

\begin{equation}
    \alpha_{\hat{i}+1}
=
\alpha_{\hat{i}}
+
\gamma\left(
\alpha
-
\sum_{\hat{s}=1}^{\hat{i}} w_{\hat{s}} \,\mathrm{err}_{\hat{s}}
\right),
\label{aci_rule}
\end{equation}
where 
\[
w_{\hat{s}}
=
\frac{0.95^{\hat{i}-\hat{s}}}{\displaystyle\sum_{k=1}^{\hat{i}}0.95^{\hat{i}-k}},
\qquad 1\leq \hat{s}\leq \hat{i} .
\]
The proposed LACI process improves the ACI process (\ref{aci_rule}) using only the different weighting rule 
\begin{equation}
    \widetilde{w}_{\hat{s}}
=
\frac{0.95^{\hat{i}-\hat{s}} \left(LE_{\hat{s}}^{(L)}\right)}{\displaystyle\sum_{k=1}^{\hat{i}}0.95^{\hat{i}-k}\left(LE_k^{(L)}\right)},
\qquad 1\leq \hat{s}\leq \hat{i}
\end{equation}

where ${LE_{\hat{s}}^{L}} = \max\big((1 - \alpha) - {LPICP}_{\hat{s}}^{L},\, 0\big)$ is the error obtained with the ACI process. Further, we compute

\[
\frac{\widetilde{w}_{\hat{s}}}{w_{\hat{s}}}
=
\frac{
0.95^{\hat{i}-\hat{s}}\left(LE_{\hat{s}}^{(L)}\right)
}{
\displaystyle
\sum_{k=1}^{\hat{i}}
0.95^{\hat{i}-k}\left(LE_k^{(L)}\right)
}
\times
\frac{
\displaystyle\sum_{k=1}^{\hat{i}}0.95^{\hat{i}-k}
}{
0.95^{\hat{i}-\hat{s}}
}
\]

that gives
\begin{equation}
    \frac{\widetilde{w}_{\hat{s}}}{w_{\hat{s}}}
=
\frac{LE_{\hat{s}}^{(L)}}{\overline{LE}_{\hat{i}}^{(L)}} \label{ws}
\end{equation}
given that \[
\overline{LE}_{\hat{i}}^{(L)}
=
\frac{
\displaystyle\sum_{k=1}^{\hat{i}}
0.95^{\hat{i}-k}LE_k^{(L)}
}{
\displaystyle\sum_{k=1}^{\hat{i}}0.95^{\hat{i}-k}
}.
\]
Noting that  \(\mathrm{LE}_{\hat{s}}^{(L)} > \overline{\mathrm{LE}}_{\hat{i}}\)  leads \(\widetilde{w}_{\hat{s}} > w_{\hat{s}}\) in view of (\ref{ws}).

\subsection{Proof of Proposition 2}

\begin{proof}

Consider the ACI and LACI recursion  starting with the initial miscoverage $\alpha_1$ till the $\hat{i}^{th}$ time index
\begin{align*}
\alpha_{\hat{i}+1}^{\mathrm{ACI}}
&=
\alpha_{\hat{i}}+
\gamma\left(
\alpha-\widehat e_{\hat{i}}^{\mathrm{ACI}}
\right),\\
 \mbox{and~} \alpha_{\hat{i}+1}^{\mathrm{LACI}}
&=
\alpha_{\hat{i}}+
\gamma\left(
\alpha-\widehat e_{\hat{i}}^{\mathrm{LACI}}
\right),
\end{align*}

where 
\[
\widehat e_{\hat{i}}^{\mathrm{ACI}}
=
\sum_{\hat{s}=1}^{\hat{i}} w_{\hat{s}}\,\mathrm{err}_{\hat{s}}
\quad \mbox{and} \quad
\widehat e_{\hat{i}}^{\mathrm{LACI}}
=
\sum_{\hat{s}=1}^{\hat{i}} \widetilde{w}_{\hat{s}}\,\mathrm{err}_{\hat{s}}.
\] respectively. Further, $\{LE_k\}_{k=1}^{\hat{i}}$ be the trajectory of the local coverage error obtained by the ACI recursion satisfying the condition (\ref{ass:severity-dominance}).

Now using (\ref{ws}), we compute
\begin{align*}
\widehat e_{\hat{i}}^{\mathrm{LACI}}
-\widehat e_{\hat{i}}^{\mathrm{ACI}}
=
\sum \limits_{\hat{s}\in\mathcal{F}_{\hat{i}}}
\left(\widetilde w_{\hat{s}}-w_{\hat{s}}\right)
=
\frac{
\sum_{\hat{s}\in\mathcal{F}_{\hat{i}}}
w_{\hat{s}}\left(
LE_{\hat{s}}^{(L)}-\overline{LE}_{\hat{i}}^{(L)}
\right)
}{
\overline{LE}_{\hat{i}}^{(L)}
} .
\end{align*}

As \(p\to 1\) in (\ref{ass:severity-dominance}), the following quantity is strictly positive:
\[
\sum_{\hat{s}\in\mathcal{F}_{\hat{i}}}
w_{\hat{s}}\left(
LE_{\hat{s}}^{(L)}-\overline{LE}_{\hat{i}}^{(L)}
\right)>0.
\]

This leads to 
\begin{equation}
\label{eq:prop2_error_order}
\widehat e_{\hat{i}}^{\mathrm{LACI}}
>
\widehat e_{\hat{i}}^{\mathrm{ACI}}.
\end{equation}

Further,
$
\alpha_{\hat{i}+1}^{\mathrm{LACI}}
-
\alpha_{\hat{i}+1}^{\mathrm{ACI}}
=
-\gamma
\left(
\widehat e_{\hat{i}}^{\mathrm{LACI}}
-
\widehat e_{\hat{i}}^{\mathrm{ACI}}
\right)
$
with  $\gamma>0$, follows
\begin{equation}
\label{eq:prop2_alpha_order}
\alpha_{\hat{i}+1}^{\mathrm{LACI}}
<
\alpha_{\hat{i}+1}^{\mathrm{ACI}}.
\end{equation}

In other words, the LACI produces a stronger corrective
reduction in the miscoverage level than
recency-weighted ACI in case of abrupt changes in auto-regressive distribution. It makes the LACI obtains more larger prediction set than the ACI in such cases leading to 
\begin{equation}
z{(h_{{i}+1})}^{\mathrm{LACI}} >  z{(h_{{i}+1})}^{\mathrm{ACI}}  \label{zh}    
\end{equation}
where $z(h_{{i}+1}) := \mathbb{I}\{x_{{i}+2} \in \hat{C}^{(\alpha_{{i}})} (h_{{i}})\}$.
Summing the (\ref{zh}) across the window  $\mathcal{N}_w^{L}(h_{{i}+1})$  gives 

\[
LPICP_{\hat{i}+1}^{(L),\mathrm{LACI}}
>
LPICP_{\hat{i}+1}^{(L),\mathrm{ACI}}
\qquad
and 
\qquad
LE_{\hat{i}+1}^{(L),\mathrm{LACI}}
<
LE_{\hat{i}+1}^{(L),\mathrm{ACI}}.
\]





\end{proof}

\subsection{Weighted ACI and GCR}
\label{appendix:weighted_aci_gcr}

\begin{proposition}[Global calibration of weighted ACI]
\label{prop:weighted_aci_gcr}
Let $\alpha \in (0,1)$, $\gamma > 0$ and $M \in (0,1)$, and consider the
recency-weighted ACI process
\begin{equation*}
  \alpha_{\hat{i}+1} = \alpha_{\hat{i}} + \gamma\Bigl(\alpha - \sum_{\hat{s}=1}^{\hat{i}} w_{\hat{s}}\,\mathrm{err}_{\hat{s}}\Bigr),
  \qquad
  w_{\hat{s}} = \frac{M^{\,\hat{i}-\hat{s}}}{\sum_{\hat{s}'=1}^{\hat{i}} M^{\,\hat{i}-\hat{s}'}},
\end{equation*}
applied to a sequence of time series observations $\{x_{\hat{i}}\}_{\hat{i}=1}^{t}$, with
$\alpha_1 \in [0,1]$ and prediction sets satisfying
$\widehat C^{\alpha_{\hat{i}}}(h_{\hat{i}}) = \mathbb{R}$ if $\alpha_{\hat{i}} < 0$ and
$\widehat C^{\alpha_{\hat{i}}}(h_{\hat{i}}) = \varnothing$ if $\alpha_{\hat{i}} > 1$. Then, for any
sequence of observations,
\begin{equation*}
  \lim_{t\to\infty} \frac{1}{t} \sum_{\hat{i}=1}^{t} \mathrm{err}_{\hat{i}} = \alpha .
\end{equation*}
\end{proposition}

\begin{proof}

Recall the recency-weighted ACI update rule:
\begin{equation}
\alpha_{\hat{i}+1}
=
\alpha_{\hat{i}}
+
\gamma\left(
\alpha-\sum_{\hat{s}=1}^{\hat{i}}w_{\hat{s}}\mathrm{err}_{\hat{s}}
\right),
\label{eq:weighted_aci_gcr_update}
\end{equation}

Let the weighted average of historical
coverage errors at time $\hat{i}$ be denoted by
\[
\widehat e_{\hat{i}}
=
\sum_{\hat{s}=1}^{\hat{i}}w_{\hat{s}}\mathrm{err}_{\hat{s}}.
\]

Rearranging \eqref{eq:weighted_aci_gcr_update} gives
$\widehat e_{\hat{i}} = \alpha - (\alpha_{\hat{i}+1}-\alpha_{\hat{i}})/\gamma$.
Summing over $\hat{i} = 1,\ldots,t$, the right-hand side telescopes, so that
$\sum_{\hat{i}=1}^{t} \widehat e_{\hat{i}} = t\alpha - (\alpha_{t+1}-\alpha_1)/\gamma$.
Therefore,
\begin{equation}
\frac{1}{t}
\sum_{\hat{i}=1}^{t}\widehat e_{\hat{i}}-\alpha
=
-\frac{\alpha_{t+1}-\alpha_1}{\gamma t}.
\label{eq:gcr_weighted_average}
\end{equation}

\medskip
\noindent

For each time $\hat{i}$, we can write
\[
\mathrm{err}_{\hat{i}}
=
\widehat e_{\hat{i}}+
(\mathrm{err}_{\hat{i}}-\widehat e_{\hat{i}}).
\]

Summing and dividing by $t$,
\[
\frac{1}{t}\sum_{\hat{i}=1}^{t}\mathrm{err}_{\hat{i}}
=
\frac{1}{t}\sum_{\hat{i}=1}^{t}\widehat e_{\hat{i}}
+
\frac{1}{t}\sum_{\hat{i}=1}^{t}
(\mathrm{err}_{\hat{i}}-\widehat e_{\hat{i}}).
\]

that gives in view of (\ref{eq:gcr_weighted_average}),
\begin{equation}
\frac{1}{t}\sum_{\hat{i}=1}^{t}\mathrm{err}_{\hat{i}}-\alpha
=
-\frac{\alpha_{t+1}-\alpha_1}{\gamma t}
+
\frac{1}{t}\sum_{\hat{i}=1}^{t}
(\mathrm{err}_{\hat{i}}-\widehat e_{\hat{i}}).
\label{eq:gcr_error_decomposition}
\end{equation}

Further let us rewrite
\[
\widehat e_{\hat{i}}=\frac{A_{\hat{i}}}{B_{\hat{i}}} \qquad  \mbox{with~~~}  A_{\hat{i}}=\sum_{\hat{s}=1}^{\hat{i}}M^{\hat{i}-\hat{s}}\mathrm{err}_{\hat{s}}
\mbox{~~and~~}
B_{\hat{i}}=\sum_{\hat{s}=1}^{\hat{i}}M^{\hat{i}-\hat{s}}.
\]
that satisfies the recursive relations
\[
A_{\hat{i}}=MA_{\hat{i}-1}+\mathrm{err}_{\hat{i}},
\mbox{~~~and~~~}
B_{\hat{i}}=MB_{\hat{i}-1}+1,
\]
with $A_0=B_0=0$.

Using $A_{\hat{i}}=B_{\hat{i}}\widehat e_{\hat{i}}$,
we obtain
\[
B_{\hat{i}}\widehat e_{\hat{i}}
=
MB_{\hat{i}-1}\widehat e_{\hat{i}-1}
+\mathrm{err}_{\hat{i}}.
\]

Since $B_{\hat{i}}=MB_{\hat{i}-1}+1$,
\[
(MB_{\hat{i}-1}+1)\widehat e_{\hat{i}}
=
MB_{\hat{i}-1}\widehat e_{\hat{i}-1}
+\mathrm{err}_{\hat{i}}.
\]

Rearranging,
\begin{equation}
\mathrm{err}_{\hat{i}}-\widehat e_{\hat{i}}
=
\kappa_{\hat{i}}
(\widehat e_{\hat{i}}-\widehat e_{\hat{i}-1}),
\label{eq:gcr_filter_identity}
\end{equation}
where
\[
\kappa_{\hat{i}}=MB_{\hat{i}-1}
=
\frac{M(1-M^{\hat{i}-1})}{1-M}.
\]

Noting that $\kappa_{\hat{i}}$ is non-decreasing with $\kappa_1=0$ and
\[
0\leq\kappa_{\hat{i}}<\frac{M}{1-M},~~ \hat{i}=1,2,\ldots t,
\]

Summing (\ref{eq:gcr_filter_identity})
over $\hat{i}=1,\ldots,t$,
\[
\sum_{\hat{i}=1}^{t}
(\mathrm{err}_{\hat{i}}-\widehat e_{\hat{i}})
=
\sum_{\hat{i}=1}^{t}
\kappa_{\hat{i}}(\widehat e_{\hat{i}}-\widehat e_{\hat{i}-1}).
\]

Expanding and rearranging the sums leads
\begin{align*}
&\sum_{\hat{i}=1}^{t}
(\mathrm{err}_{\hat{i}}-\widehat e_{\hat{i}})
=
\kappa_t\widehat e_t
-
\sum_{\hat{i}=1}^{t-1}
(\kappa_{\hat{i}+1}-\kappa_{\hat{i}})\widehat e_{\hat{i}}.
\end{align*}

Since $0\leq\widehat e_{\hat{i}}\leq1$ and
$\kappa_{\hat{i}+1}-\kappa_{\hat{i}}\geq0$,
we have
\[
0\leq\kappa_t\widehat e_t\leq\kappa_t
\]
and
\[
0\leq
\sum_{\hat{i}=1}^{t-1}
(\kappa_{\hat{i}+1}-\kappa_{\hat{i}})\widehat e_{\hat{i}}
\leq
\kappa_t-\kappa_1=\kappa_t.
\]

Therefore,
\[
\left|
\sum_{\hat{i}=1}^{t}
(\mathrm{err}_{\hat{i}}-\widehat e_{\hat{i}})
\right|
\leq\kappa_t
<
\frac{M}{1-M}.
\]

Dividing by $t$ gives
\begin{equation}
\boxed{
\left|
\frac{1}{t}\sum_{\hat{i}=1}^{t}
(\mathrm{err}_{\hat{i}}-\widehat e_{\hat{i}})
\right|
\leq
\frac{M}{(1-M)t}.
}
\label{eq:gcr_smoothing_bound}
\end{equation}

Since $M$ is fixed with $0<M<1$,
the right-hand side approaches zero
as $t\to\infty$.

From (\ref{eq:gcr_error_decomposition}), we obtain
\begin{equation}
\left| \frac{1}{t}\sum_{\hat{i}=1}^{t}\mathrm{err}_{\hat{i}}-\alpha \right|
\leq 
\left| \frac{\alpha_{t+1}-\alpha_1}{\gamma t} \right|
+
\left| \frac{1}{t}\sum_{\hat{i}=1}^{t}
(\mathrm{err}_{\hat{i}}-\widehat e_{\hat{i}}) \right|. \label{45}
\end{equation}

It remains to bound the first term of above inequality.
Adapting the argument of \citet[Lemma~4.1]{gibbs2021adaptive}, once
$\alpha_{\hat{i}}$ leaves $[0,1]$, the boundary convention forces
$\mathrm{err}_{\hat{i}} = 0$ (if $\alpha_{\hat{i}} < 0$) or $\mathrm{err}_{\hat{i}} = 1$
(if $\alpha_{\hat{i}} > 1$). Due to the exponential smoothing, $\widehat e_{\hat{i}}$ then
crosses $\alpha$ within at most
$K_{\mathrm{lo}} = \lceil \log\alpha / \log M \rceil$ or
$K_{\mathrm{up}} = \lceil \log(1-\alpha) / \log M \rceil$ steps, respectively,
after which $\alpha_{\hat{i}}$ returns towards $[0,1]$. Hence, for all $\hat{i}$,
\[
  -\gamma(1-\alpha)K_{\mathrm{lo}} \;\le\; \alpha_{\hat{i}} \;\le\; 1 + \gamma\alpha K_{\mathrm{up}} .
\]
Since $\alpha_1 \in [0,1]$, the distance between the initial and final levels
is at most the larger of $1 + \gamma\alpha K_{\mathrm{up}}$ (attained when
$\alpha_1 = 0$) and $1 + \gamma(1-\alpha)K_{\mathrm{lo}}$ (attained when
$\alpha_1 = 1$), that is,
\[
  |\alpha_{t+1} - \alpha_1| \;\le\; D
  := 1 + \gamma \max\bigl\{(1-\alpha)K_{\mathrm{lo}},\; \alpha K_{\mathrm{up}}\bigr\}.
\]
The constant $D$ depends only on $\alpha$, $\gamma$ and $M$, and not on $t$.
Therefore,
\begin{equation}
  \left| \frac{\alpha_{t+1}-\alpha_1}{\gamma t} \right| \;\le\; \frac{D}{\gamma t}
  \;\longrightarrow\; 0 \quad (t \to \infty).
  \label{eq:gcr_drift_bound}
\end{equation}

Using (\ref{eq:gcr_smoothing_bound})
and (\ref{eq:gcr_drift_bound}) in (\ref{45}),
we obtain
\begin{equation}
\boxed{
\left|
\frac{1}{t}\sum_{\hat{i}=1}^{t}
\mathrm{err}_{\hat{i}}-\alpha
\right|
\leq
\frac{D}{\gamma t}
+
\frac{M}{(1-M)t}.
}
\label{eq:gcr_weighted_aci_bound}
\end{equation}

Since $D$, $\gamma$, and $M$
are fixed, both terms on the
right-hand side converge to zero
as $t\to\infty$.

Therefore,
\begin{equation}
\boxed{
\lim_{t\to\infty}
\frac{1}{t}
\sum_{\hat{i}=1}^{t}\mathrm{err}_{\hat{i}}
=
\alpha.
}
\label{eq:gcr_weighted_aci_result}
\end{equation}

\end{proof}

\section{ Additional Experiments with other Neural Architectures and Tunning Details}

\begin{table}[H]
\centering
\vspace{-2mm}

{\fontsize{8.2pt}{8.2pt}\selectfont
\setlength{\tabcolsep}{6.8pt}
\renewcommand{\arraystretch}{0.85}

\begin{tabular}{llp{6.5cm}}
\toprule
\textbf{Category} & \textbf{Hyperparameter} & \textbf{Value} \\
\midrule

\multirow{7}{*}{Global Training}
& Sequence length (SEQ\_LEN) & 48 \\
& Initial training epochs (EPOCHS) & 200 \\
& Online retraining & 1 epoch every 6 steps \\
& Learning rate (LR) & 0.001 \\
& Optimizer & Adam \\
& Random seed (SEED) & 42 \\
& Batch size (BATCH\_SIZE) & 32 \\

\midrule

\multirow{4}{*}{Conformal Prediction}
& Miscoverage rate ($\alpha$) & 0.1 \\
& Initial window size (TINIT) & 5000 \\
& Train/calibration split (SPLIT) & 0.5 \\
& Adaptation step size ($\gamma$, ACI/LACI) & 0.001 \\

\midrule

\multirow{3}{*}{LACI}
& Local window size ($w$) & 24 \\
& Exponential decay factor & 0.85 \\
& Weighting epsilon ($\varepsilon$) & $10^{-4}$ \\

\midrule

\multirow{2}{*}{DtACI}
& Gamma candidates & \{0.001, 0.002, 0.004, 0.008, 0.016\} \\
& Interval size & 500 \\

\midrule

\multirow{2}{*}{AgACI}
& Gamma candidates & \{0.001, 0.002, 0.004, 0.008, 0.016\} \\
& Epsilon (EPS) & 0.001 \\

\midrule

\multirow{5}{*}{TCN Architecture}
& Input dimension & 1 \\
& Number of channels & [32, 32, 32, 32] \\
& Kernel size & 2 \\
& Dropout & 0.2 \\
& Output head & Linear(32, 1) \\

\midrule

\multirow{5}{*}{LSTM Architecture}
& Input dimension & 1 \\
& Hidden dimension & 64 \\
& Number of layers & 2 \\
& Dropout & 0.1 \\
& Output head & Linear(64, 1) \\

\midrule

\multirow{4}{*}{GRU Architecture}
& Input dimension & 1 \\
& Hidden dimension & 64 \\
& Number of layers & 2 \\
& Output head & Linear(64, 1) \\

\bottomrule
\end{tabular}
}

\caption{Hyperparameters used in the experimental setup
for ACI, LACI, DtACI, and AgACI across TCN, LSTM,
and GRU forecasting architectures.A small numerical stabilization constant \(\epsilon=10^{-4}\) is added to  to avoid division by zero.}

\label{tab:hyperparams}
\vspace{-2mm}
\end{table}

\section{Empirical Verification of the Local-Error Assumption across Different ACI Settings}
\label{app:local_error_assumption_different_ACI}

We further empirically verify the assumption at coverage failures for different values of $\gamma$. 
To examine the robustness of this relationship, we evaluate the assumption at $\gamma \in \{0.004, 0.008, 0.016\}$. The results are reported in Tables~\ref{tab:local_error_gamma_004}--\ref{tab:local_error_gamma_016}.

\begin{table}[H]
\centering
\begin{tabular}{lrrrrrr}
\hline
Dataset
& $\#(err_{\hat{s}}=1)$
& $\#(LE_{\hat{s}} > \overline{LE}_{\hat{s}} \mid err_{\hat{s}}=1)$
& (\%)
& Mean $LE_{\hat{s}}^{(L)}$
& Mean $\overline{LE}_{\hat{s}}^{(L)}$
& $\Delta$ \\
\hline

50Hertz
& 409 & 259 & 63.33 & 0.08 & 0.04 & 0.04 \\

Amprion
& 401 & 247 & 61.60 & 0.11 & 0.06 & 0.04 \\

Location1
& 476 & 272 & 57.14 & 0.08 & 0.04 & 0.04 \\

Location2
& 439 & 241 & 54.90 & 0.06 & 0.04 & 0.03 \\

Plant 1 Generation
& 198 & 127 & 64.14 & 0.09 & 0.07 & 0.03 \\

Plant 2 Generation
& 348 & 272 & 78.16 & 0.10 & 0.07 & 0.03 \\

TenneTTSO
& 420 & 258 & 61.43 & 0.10 & 0.06 & 0.04 \\

TransnetBW
& 434 & 251 & 57.83 & 0.06 & 0.03 & 0.03 \\

Hydropower
& 149 & 105 & 70.47 & 0.16 & 0.08 & 0.07 \\

\hline
\textbf{Total}
& \textbf{3274}
& \textbf{2032}
& \textbf{62.06}
& \textbf{0.09}
& \textbf{0.05}
& \textbf{0.04} \\

\hline
\end{tabular}

\caption{Local coverage errors of ACI with $\gamma=0.004$
across nine energy datasets. The table reports the number
of eligible coverage-failure observations for which
$\mathrm{LE}_{\hat{s}}^{(L)} >
\overline{\mathrm{LE}}_{\hat{s}}^{(L)}$.
Across all datasets, this condition holds for
2032 out of 3274 eligible coverage failures (62.06\%).
The reported mean local errors and weighted-average
baselines are computed at eligible coverage-failure
observations. The final row reports the pooled proportion
and the pooled mean difference $\Delta$.}

\label{tab:local_error_gamma_004}
\end{table}

\begin{table}[H]
\centering
\begin{tabular}{lrrrrrr}
\hline
Dataset
& $\#(err_{\hat{s}}=1)$
& $\#(LE_{\hat{s}} > \overline{LE}_{\hat{s}} \mid err_{\hat{s}}=1)$
& (\%)
& Mean $LE_{\hat{s}}^{(L)}$
& Mean $\overline{LE}_{\hat{s}}^{(L)}$
& $\Delta$ \\
\hline

50Hertz
& 430 & 263 & 61.16 & 0.08 & 0.04 & 0.04 \\

Amprion
& 417 & 266 & 63.79 & 0.10 & 0.05 & 0.04 \\

Location1
& 464 & 263 & 56.68 & 0.07 & 0.04 & 0.04 \\

Location2
& 460 & 259 & 56.30 & 0.06 & 0.03 & 0.03 \\

Plant 1 Generation
& 259 & 183 & 70.66 & 0.07 & 0.04 & 0.03 \\

Plant 2 Generation
& 366 & 284 & 77.60 & 0.10 & 0.07 & 0.03 \\

TenneTTSO
& 435 & 257 & 59.08 & 0.09 & 0.05 & 0.04 \\

TransnetBW
& 438 & 259 & 59.13 & 0.06 & 0.03 & 0.03 \\

Hydropower
& 147 & 96 & 65.31 & 0.13 & 0.07 & 0.06 \\

\hline
\textbf{Total}
& \textbf{3416}
& \textbf{2130}
& \textbf{62.35}
& \textbf{0.08}
& \textbf{0.05}
& \textbf{0.03} \\

\hline
\end{tabular}

\caption{Local coverage errors of ACI with $\gamma=0.008$  across nine energy
datasets. The table reports the number of eligible
coverage-failure observations for which
$\mathrm{LE}_{\hat{s}}^{(L)} >
\overline{\mathrm{LE}}_{\hat{s}}^{(L)}$.
Across all datasets, this condition holds for
2130 out of 3416 eligible coverage failures (62.35\%).
The reported mean local errors and weighted-average
baselines are computed at eligible coverage-failure
observations. The final row reports the pooled proportion
and pooled mean difference $\Delta$.}

\label{tab:local_error_gamma_008}
\end{table}

\begin{table}[H]
\centering
\begin{tabular}{lrrrrrr}
\hline
Dataset
& $\#(err_{\hat{s}}=1)$
& $\#(LE_{\hat{s}} > \overline{LE}_{\hat{s}} \mid err_{\hat{s}}=1)$
& (\%)
& Mean $LE_{\hat{s}}^{(L)}$
& Mean $\overline{LE}_{\hat{s}}^{(L)}$
& $\Delta$ \\
\hline

50Hertz
& 438 & 263 & 60.05 & 0.07 & 0.03 & 0.04 \\

Amprion
& 433 & 269 & 62.12 & 0.08 & 0.04 & 0.04 \\

Location1
& 467 & 264 & 56.53 & 0.06 & 0.03 & 0.03 \\

Location2
& 467 & 257 & 55.03 & 0.05 & 0.03 & 0.02 \\

Plant 1 Generation
& 288 & 219 & 76.04 & 0.05 & 0.03 & 0.03 \\

Plant 2 Generation
& 360 & 286 & 79.44 & 0.07 & 0.04 & 0.03 \\

TenneTTSO
& 437 & 260 & 59.50 & 0.07 & 0.04 & 0.03 \\

TransnetBW
& 444 & 254 & 57.21 & 0.05 & 0.03 & 0.02 \\

Hydropower
& 148 & 92 & 62.16 & 0.10 & 0.05 & 0.05 \\

\hline
\textbf{Total}
& \textbf{3482}
& \textbf{2164}
& \textbf{62.15}
& \textbf{0.07}
& \textbf{0.04}
& \textbf{0.03} \\

\hline
\end{tabular}

\caption{Local coverage errors with $\gamma=0.016$   of ACI across nine energy
datasets. The table reports the number of eligible
coverage-failure observations for which
$\mathrm{LE}_{\hat{s}}^{(L)} >
\overline{\mathrm{LE}}_{\hat{s}}^{(L)}$.
Across all datasets, this condition holds for
2164 out of 3482 eligible coverage failures (62.15\%).
The reported mean local errors and weighted-average
baselines are computed at eligible coverage-failure
observations. The final row reports the pooled proportion
and pooled mean difference $\Delta$.}

\label{tab:local_error_gamma_016}
\end{table}
These results provide consistent empirical support for the local-error assumption over a range of $\gamma$ values. In particular, at coverage-failure points, the local error is systematically larger than the overall error on the majority of instances, with a positive mean difference for every value of $\gamma$ considered. This suggests that recent local miscalibration contains useful information about imminent coverage failures, providing further empirical justification for incorporating local error information into the proposed calibration procedure.

\end{document}